\documentclass[sigconf, screen=true, printacmref=false, printccs=false, printfolios=false]{acmart}
\setcopyright{none}

\usepackage{xcolor}
\definecolor{thedarkblue}{RGB}{0,0,120} 
\definecolor{mydarkblue}{rgb}{0,0.08,0.45} 
\definecolor{darkblue}{rgb}{0,0.08,180}
\colorlet{TufteRed}{red!80!black}
\definecolor{theblue}{RGB}{0,0,180}
\colorlet{thered}{TufteRed}

\usepackage{microtype}
\usepackage{balance}

\usepackage{booktabs}
\usepackage{tabularx}

\usepackage{amsmath,amsthm}

\usepackage{tikz}
\usepackage{verbatim}
\usetikzlibrary{arrows}
\usetikzlibrary{shapes,snakes}
\usetikzlibrary{decorations.pathmorphing} 
\usetikzlibrary{fit}					
\usetikzlibrary{backgrounds}	

\usepackage{ragged2e}
\usepackage{multirow}
\usepackage{microtype}
\usepackage{balance}
\usepackage{setspace}

\graphicspath{{./}{./graphics/}}

\newcolumntype{R}[1]{>{\RaggedLeft\arraybackslash}} 
\newcolumntype{L}[1]{>{\RaggedRight\arraybackslash}}

\newcommand{\eg}{\emph{e.g.}}
\newcommand{\ie}{\emph{i.e.}}

\newtheorem{Definition}{\hspace{-1em}\bfseries{Definition}}

\AtBeginEnvironment{pmatrix}{\setlength{\arraycolsep}{2pt}}

\providecommand{\mat}[1]{\boldsymbol{\mathrm{#1}}}
\renewcommand{\vec}[1]{\boldsymbol{\mathrm{#1}}}

\DeclareMathOperator{\hugeE}{\mbox{\huge\raise-0.3ex\hbox{E}}}
\DeclareMathOperator{\p}{\mathbb{P}}
\DeclareMathOperator{\hugep}{\mbox{\huge\raise-0.3ex\hbox{$\p$}}}

\newcommand{\RR}{\mathbb{R}}

\providecommand{\mA}{\ensuremath{\mat{A}}}

\providecommand{\mD}{\ensuremath{\mat{D}}}

\providecommand{\mH}{\ensuremath{\mat{H}}}

\providecommand{\mP}{\ensuremath{\mat{P}}}

\providecommand{\mW}{\ensuremath{\mat{W}}}
\providecommand{\mX}{\ensuremath{\mat{X}}}
\providecommand{\mY}{\ensuremath{\mat{Y}}}
\providecommand{\mZ}{\ensuremath{\mat{Z}}}

\providecommand{\vc}{\ensuremath{\vec{c}}}
\providecommand{\vd}{\ensuremath{\vec{d}}}

\providecommand{\vw}{\ensuremath{\vec{w}}}
\providecommand{\vx}{\ensuremath{\vec{x}}}
\providecommand{\vy}{\ensuremath{\vec{y}}}
\providecommand{\vz}{\ensuremath{\vec{z}}}

\usepackage{subfigure}
\usepackage{graphbox}
\usepackage{svg}
\usepackage{nicefrac}

\DeclareMathAlphabet{\mathbcal}{OMS}{cmsy}{b}{n}
\usepackage{amsmath}
\usepackage{comment}

\usepackage{bm}
\usepackage{bbm}
\usepackage{enumitem}
\AtBeginDocument{
\providecommand\BibTeX{{
\normalfont B\kern-0.5em{\scshape i\kern-0.25em b}\kern-0.8em\TeX}}}

\usepackage{microtype}

\usepackage{algorithm}
\usepackage{algorithmicx}
\usepackage{algpseudocode}
\algblockdefx[parallel]{ParFor}{EndPar}[1][]{$\textbf{parallel for}$ #1 $\textbf{do}$}{$\textbf{end parallel}$}
\algrenewcommand{\alglinenumber}[1]{\fontsize{6.5}{7}\selectfont#1}
\algtext*{EndPar}

\algblockdefx[parallel]{parfor}{endpar}[1][]{$\textbf{parallel for}$ #1 $\textbf{do}$}{$\textbf{end parallel}$}
\algrenewcommand{\alglinenumber}[1]{\scriptsize#1:}

\algblockdefx[parallel]{parallelfor}{parallelend}
[1][]{\textbf{parallel for} #1}
{\textbf{end parallel}}

\usepackage{nicefrac}

\algnotext{EndIf}
\algnotext{EndProcedure}

\usepackage{xargs}      
\usepackage{color}

\definecolor{lightpink}{RGB}{237,157,202}
\definecolor{lightred}{RGB}{210,121,121}
\definecolor{lightorange}{RGB}{230,170,50}
\definecolor{lightgold}{RGB}{210,194,121}
\definecolor{lightgreen}{RGB}{121,210,121}
\definecolor{lightaqua}{RGB}{121,206,210}
\definecolor{lightblue}{RGB}{121,124,210}
\definecolor{lightpurple}{RGB}{153,102,255}
\definecolor{red}{RGB}{178,34,34}
\definecolor{gray}{RGB}{166,166,166}

\definecolor{mydarkblue}{rgb}{0,0.08,0.45} 
\usepackage{hyperref}
\hypersetup{%
    colorlinks=true,
    linkcolor=mydarkblue,
    citecolor=mydarkblue,
    filecolor=mydarkblue,
    urlcolor=mydarkblue}

\begin{document}

\title{A Hypergraph Neural Network Framework for Learning Hyperedge-Dependent Node Embeddings}

\settopmatter{authorsperrow=5}

\author{Ryan Aponte}
\affiliation{
\institution{CMU}
}

\author{Ryan A. Rossi}
\affiliation{
\institution{Adobe Research}
}

\author{Shunan Guo}
\affiliation{
\institution{Adobe Research}
}

\author{Jane Hoffswell}
\affiliation{
\institution{Adobe Research}
}

\author{Nedim Lipka}
\affiliation{
\institution{Adobe Research}
}

\author{Chang Xiao}
\affiliation{
\institution{Adobe Research}
}

\author{Gromit Chan}
\affiliation{
\institution{Adobe Research}
}

\author{Eunyee Koh}
\affiliation{
\institution{Adobe Research}
}

\author{Nesreen Ahmed}
\affiliation{
\institution{Intel Labs}
}
\email{}

\renewcommand{\shortauthors}{R. Aponte et al.}

\begin{abstract} 
In this work, we introduce a hypergraph representation learning framework called Hypergraph Neural Networks (HNN) that jointly learns hyperedge embeddings along with a set of hyperedge-dependent embeddings for each node in the hypergraph. HNN derives multiple embeddings per node in the hypergraph where each embedding for a node is dependent on a specific hyperedge of that node. Notably, HNN is accurate, data-efficient, flexible with many interchangeable components, and useful for a wide range of hypergraph learning tasks. We evaluate the effectiveness of the HNN framework for hyperedge prediction and hypergraph node classification. 
We find that HNN achieves an overall mean gain of 7.72\% and 11.37\% across all baseline models and graphs for hyperedge prediction and hypergraph node classification, respectively. 
\end{abstract}



\maketitle

\section{Introduction}
In the real world, it is common for a single relationship to involve more than two entities (e.g., multiple authors collaborating on a paper that cites a body of related work~\cite{DBLP:journals/corr/abs-1205-6233,sen:aimag08}), whereas prior research on graph neural networks (GNNs) only permits a maximum of two entities per relationship~\cite{bastings-etal-2017-graph,berg2017graph,ying2018graph,fan2019graph, wang2019kgat,lee2018higher,he2020lightgcn,qi2018learning,chanpuriya2021interpretable,DBLP:journals/corr/abs-1901-00596}.
This oversimplification presents challenges in modeling higher-order relationships~\cite{lee2019attention,zhou2020graph,wu2020comprehensive}.
However, hypergraphs can capture these higher-order relationships by allowing edges to connect more than two nodes~\cite{thomas2022graph,li2022enhancing,sun2022motifs,kim2022equivariant,pearson2014spectral,xia2022hypergraph,di2022generating}.

Recent work has developed methods for learning node embeddings and representations from such hypergraphs~\cite{huang2021unignn,zhang2022deep,HGNN,chien2021you,hypergcn_neurips19,wu2020adahgnn}.
However, most existing approaches are based on the premise that nodes in the same hyperedge should be represented in a similar fashion.
This strict assumption is often unrealistic and violated in practice where hypergraphs are often noisy, partially observed, with missing and incomplete connections.
Most importantly, it fundamentally limits the utility of such methods for real-world applications and can often lead to poor predictive performance, especially in real-world hypergraphs where hyperedges are of any arbitrary size.
For instance, nodes typically participate in many hyperedges with non-uniform degree, and therefore, the learned embedding of a node will often be most similar to nodes in the largest hyperedge, failing to capture the other hyperedges that may often be more important to model for a downstream task.
While learning multiple node embeddings along with hyperedge embeddings can help resolve this fundamental limitation, past work has only focused on learning a single embedding per node, despite that a node often participates in many hyperedges with varying degree.

\begin{figure}[t!]
\vspace{-3mm}
\centering
\includegraphics[width=0.95\linewidth]{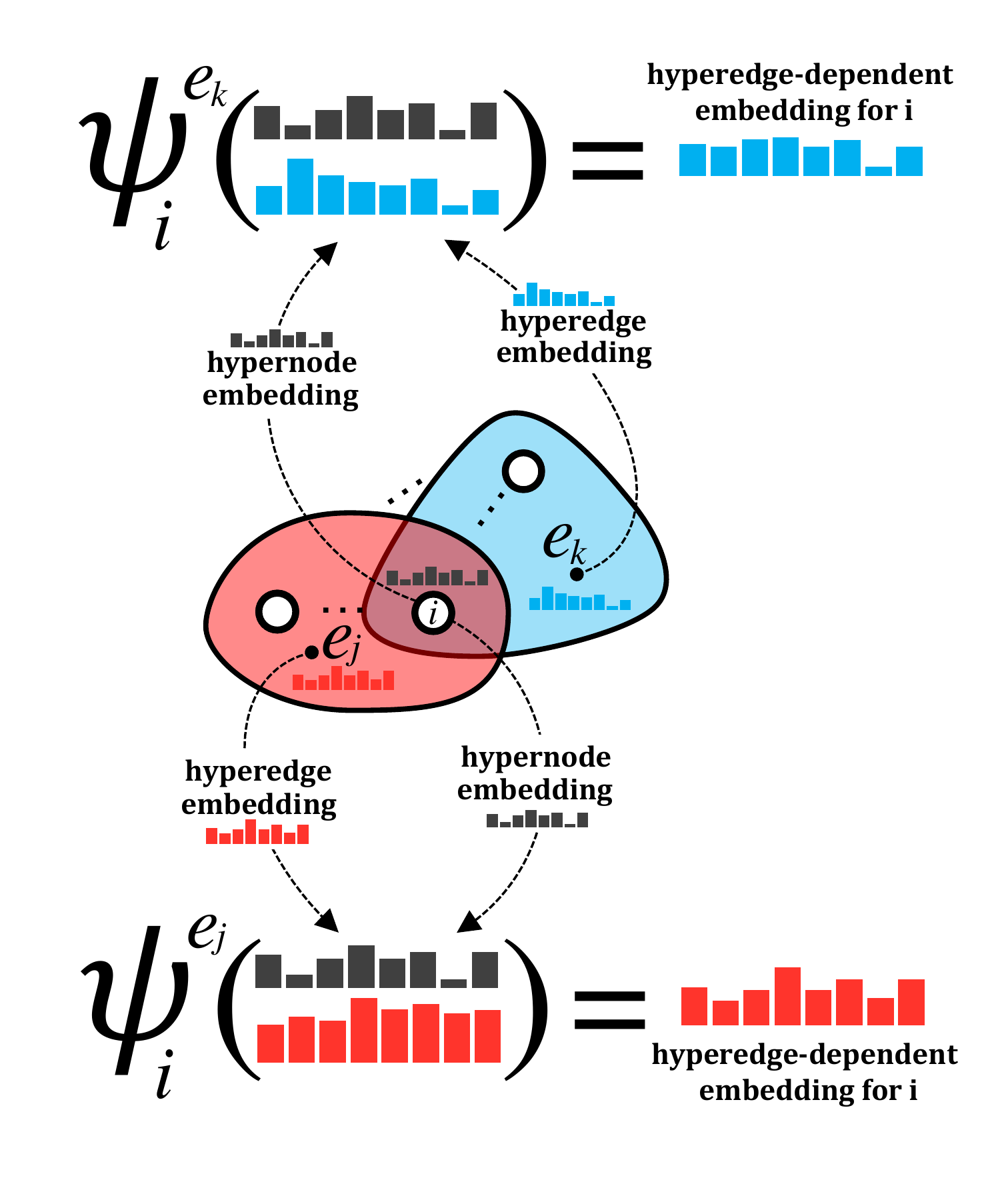}
\vspace{-10mm}
\caption{
Hyperedge-dependent embeddings for a vertex $i$.
Our approach learns multiple embeddings for vertex $i$ that are dependent on the specific hyperedge to effectively utilize higher-order relationships in the hypergraph.
}
\label{fig:hyperedge-dependent-embedding}
\vspace{-5mm}
\end{figure}

To address these issues,
we introduce a hypergraph representation learning framework called HNN that jointly learns embeddings of hyperedges as well as a set of hyperedge-dependent embeddings for each node in the hypergraph (Figure~\ref{fig:hyperedge-dependent-embedding}).
Notably, the framework explicitly learns both hyperedge embeddings and node embeddings that are optimized in an end-to-end iterative fashion.
Every node has an implicit set of embeddings associated with it called hyperedge-dependent embeddings.
The set of hyperedge-dependent embeddings of a node appropriately capture the similarity of it with respect to a specific hyperedge of interest.
Intuitively, a node $i$ may have multiple unique node embeddings depending on the precise hyperedge of interest as shown in Figure~\ref{fig:hyperedge-dependent-embedding}.
Therefore, HNN avoids the strict assumption of other approaches as well as the limitations that arise from it.
Furthermore, our approach does not require that node $i$ actually be present in the hyperedge to derive such a hyperedge-dependent embedding.
For instance, one may want to understand the similarity of a node $i$ of interest with respect to the other nodes that participate in a specific hyperedge.
Such a task can be handled in a principled fashion using HNN.
This important key advantage is 
unique to HNN and enables us to learn better contextualized embeddings that can be locally optimized for various downstream tasks like hyperlink prediction, node classification, or style recommendation.
The framework is designed to be flexible with many interchangeable components while also naturally supporting inductive learning tasks on hypergraphs such as inferring new unseen hyperedges as well as being amenable to input features on the hyperedges (as well as the nodes).
This flexibility makes it well-suited for a variety of hypergraph learning tasks such as node classification and hyperedge prediction.

To evaluate the predictive performance of our approach, we compare HNN to a variety of 
other hypergraph learning methods
for node classification, hyperedge prediction, and document style recommendation.
Notably, HNN achieves an overall mean gain of $7.72\%$ for hyperedge prediction and $11.37\%$ for hypergraph node classification across all other models and graphs.
We further evaluate HNN for design style recommendation by formulating this new task as a hypergraph representation learning problem.
For this, we derived a heterogeneous hypergraph from a large corpus of HTML documents representing marketing emails.
To make it easy for others to investigate this new style recommendation task, we release the corpus of HTML documents along with the heterogeneous hypergraph derived from it.

\medskip\noindent
A summary of the main contributions of this work are as follows:
\begin{itemize}[leftmargin=*]
\item \textbf{Problem Formulation:} 
We formulate a new hypergraph representation learning problem that aims to jointly learn hyperedge embeddings along with a set of \emph{hyperedge}\emph{-dependent} \emph{embeddings} for each node in the hypergraph.

\item \textbf{Framework:} 
We propose a hypergraph representation learning framework called HNN for this new problem. The framework is flexible with many interchangeable components, highly accurate, data-efficient, and highly effective 
for a wide variety of downstream applications.

\item \textbf{Effectiveness:}
We demonstrate the effectiveness of our approach through a comprehensive set of experiments including 
hyperedge prediction (Sec.~\ref{sec:exp-hyperedge-pred-results}), 
node classification (Sec.~\ref{sec:node-classification-results}), 
and a case study using HNN
for HTML style recommendation (Sec.~\ref{sec:email-style-recommendation}).

\item \textbf{Benchmark Data:} 
We formulate the 
HTML document style recommendation task as a hypergraph learning problem and apply our model to perform style recommendation. 
We derived a heterogeneous hypergraph from a large corpus of HTML documents representing marketing emails and release the benchmark dataset for others to use.
\end{itemize}

\section{Related Work} \label{sec:related-work}
Hypergraphs generalize relationships beyond pairwise to allow the modeling of higher-order relationships.
Hypergraph neural networks have been used for a variety of applications such as 3D pose estimation~\cite{liu2020semi}, cancer tissue classification~\cite{bakht2021colorectal}, group recommendation~\cite{jia2021hypergraph,https://doi.org/10.48550/arxiv.2103.13506}, 
friendship recommendation~\cite{zhu2021inhomogeneous}, recipe prediction~\cite{Zhang_Cui_Jiang_Chen_2018}, among many others~\cite{bai2021multi,thonet2022joint,9709542,9329123,NIPS2006_dff8e9c2,zhang2022deep,https://doi.org/10.48550/arxiv.1906.00137,imran2021ddhh,Xia_Yin_Yu_Wang_Cui_Zhang_2021,arya2020hypersage,arya2021adaptive}.
Recent work has proposed a hypergraph neural network attention model for node classification~\cite{https://doi.org/10.48550/arxiv.1901.08150}.
Hypergraphs have also been used in predicting time-series sensor network data~\cite{10.1145/3394486.3403389}. 
Satchidanand et al.\cite{10.5555/2832747.2832778} use random walks to address class imbalance. Jin et al.~\cite{10.5555/3367243.3367411} combine hypergraphs with manifolds to reduce the impact of noisy data.
Xue et al.~\cite{xue2021multiplex} applied hypergraph convolutional networks to obtain embeddings from a multiplex bipartite network.
Tudisco et al.~\cite{tudisco2021nonlinear} developed a nonlinear diffusion method for semi-supervised classification on hypergraphs.
Wang et al.~\cite{10.1145/3397271.3401133} developed a sequential recommendation approach for hypergraphs.
Feng et al.~\cite{10.1145/3178876.3186064} extend hypergraph learning to nodes with order whereas Yadati~\cite{NEURIPS2020_217eedd1} extend hypergraphs to model recursive relationships.
HyperGCN~\cite{hypergcn_neurips19} develop a method to train GCNs that are more robust to noisy data.
More recently, there has been work focusing on an equivalency between hypergraphs and undirected homogeneous graphs~\cite{zhang2022hypergraph}.
Chitra et al.~\cite{chitra2019random} showed that methods based on Laplacians derived from random walks on hypergraphs with edge-independent vertex weights do not utilize higher-order relationships in the data.
Zhang et al.~\cite{zhang2022learnable} developed an approach that modifies the hypergraph prior to learning to improve a prediction task.
While link prediction in graphs aims to infer relationships between two nodes, the goal of hyperedge prediction is to infer relationships between a set of two or more nodes~\cite{10.1145/3340531.3411870}, and thus, is fundamentally more challenging~\cite{https://doi.org/10.48550/arxiv.2106.04292,10.1145/2487788.2487802,10.1145/3206025.3206062,https://doi.org/10.48550/arxiv.1703.04823}.

A heterogeneous hypergraph is a generalization of a hypergraph where the nodes and edges may have various types associated with them~\cite{https://doi.org/10.48550/arxiv.2006.12278}.
They have been used for identifying communities~\cite{ZHENG201955} source code bug detection~\cite{https://doi.org/10.48550/arxiv.2201.12113}, traffic prediction~\cite{9766155}, among other tasks~\cite{9099985,fan2021heterogeneous,ZHU2016150,article,10.1145/3437963.3441835}.
Recent work has also studied link prediction in heterogeneous hypergraphs~\cite{fan2021heterogeneous,9099985,8198805}.
Besides heterogeneous hypergraphs, other work has focused on dynamic hypergraphs~\cite{jiang2019dynamic,zhang2018dynamic} and even signed hypergraphs~\cite{chen2020neural}.
In this paper, we introduce the design style recommendation problem for web-based HTML documents such as emails and formulate it as a heterogeneous hypergraph learning task by first deriving a heterogenous hypergraph from a large collection of curated marketing email designs from the web, then we develop an approach that learns a model from this hypergraph for recommending high quality and unified design styles.
To enable other researchers to study this new task, we release our collection of about 1000 web-based emails along with the derived heterogeneous hypergraph.

\section{Framework} \label{sec:approach}
We now describe our end-to-end hypergraph representation learning framework called HNN.
The proposed framework is flexible and consists of many interchangeable and customizable components.
We begin in Section~\ref{sec:approach-hypergraph-model} by introducing the hypergraph model, then
Section~\ref{sec:approach-hypergraph-features} describes input features to our framework and techniques for deriving them
(if not available/given as input).
In Section~\ref{sec:approach-hyperedge-dependent-convolution}, we present the hyperedge-dependent graph convolution component, and formally discuss a few HNN model variants in Section~\ref{sec:approach-variants}.
Finally, training of HNN is described in Section~\ref{sec:approach-training}.

\subsection{Hypergraph Model}\label{sec:approach-hypergraph-model}

Let $G=(V,E)$ denote a hypergraph where $V=\{v_1,\ldots,v_N\}$ are the $N=|V|$ vertices and $E=\{e_1,\ldots,e_M\} \subseteq 2^V$ is the set of $M=|E|$ hyperedges.
Hence, a hyperedge $e \in E$ is simply a set of vertices $e = \{s_1,\ldots,s_k\}$ such that $\forall s_i \in e, s_i \in V$.
Furthermore, hyperedges can be of any arbitrary size and are not restricted to a specific size, thus, $e_i, e_j \in E$, then $|e_i| < |e_j|$ may hold.
Now, we formally define a few important hypergraph notions that will be fundamental for our approach.
We first define the neighborhood of a vertex as 
$N_i = \{j \in V \,|\, e \in E \wedge i \in e \wedge j \in e \}$.
Hence, $j$ is a neighbor of $i$ iff there exists a hyperedge $e \in E$ in the hypergraph where $i \in e \wedge j \in e$, that is, both $i$ and $j$ are in the hyperedge $e$.
Now, let us define the hyperedge neighborhood of a vertex as follows:
\begin{Definition}[Hyperedge Neighborhood]\label{def:hyperedge-neighborhood}
The hyperedge neighborhood of a vertex $i$ is defined informally as the set of hyperedges that include vertex $i$.
More formally, 
\begin{align}
E_i = \{\, e \in E \;|\; i \in e  \,\}
\end{align}
\end{Definition}\noindent
Further, let $|E_i|$ be the total hyperedges in $G$ that include $i$.
Using the notion of hyperedge neighborhood (Def.~\ref{def:hyperedge-neighborhood}),
we can define a vertex neighborhood:
$N_i = \{e \in E_i \,|\, j \in e \}$.
Intuitively, $N_i = \bigcup_{e \in E_i} e$ is the set of vertices in the set of hyperedges that include $i$.

\subsubsection{Hyper-Incidence Graph}
Let $\mH$ denote the $N \times M$ hyper-incidence matrix of the hypergraph $G$ defined as:
\begin{equation} \label{eq:hypergraph-incidence-matrix}
H_{ik} = 
\begin{cases}					
1 	& \text{if } v_i \in e_k   \\
0 												& \text{otherwise} \\
\end{cases}
\end{equation}\noindent
Hence, $H_{ik}=1$ iff the vertex $v_i \in V$ is in the hyperedge $e_k \in E$ and $H_{ik}=0$ otherwise.
Intuitively, $\mH \in \RR^{N \times M}$ connects the nodes to their hyperedges and vice-versa.

\begin{Definition}\label{def:hyperedge-degree-vector}
The hyperedge degree vector $\vd^e \in \RR^{M}$ is 
\begin{align}\label{eq:hyperedge-degree-vector}
\vd^e = \mH^{\top}\mathbf{1}_{N}
\end{align}
where $\mathbf{1}_{N}$ is the $N$-dimensional vector of all ones.
Then the degree of a hyperedge $e_j \in E$ is simply $d^e_j = \sum_i H_{ij}$.
Alternatively, we can obtain degree of hyperedge $e_j$ as
\begin{align}
d^{e}_{j} = \vc_j'\mH'\mathbf{1}_{N}
\end{align}
where $\vc_j$ is a bit mask vector of all zeros but the j-th position is 1.
\end{Definition}

\subsubsection{Node Diagonal Degree Matrix}
We define the \emph{diagonal hyperedge node degree matrix} $\mD \in \RR^{N \times N}$ as
\begin{align}\label{eq:diag-hyperedge-node-degree-matrix}
\mD = \mathtt{diag}(\mH\mathbf{1}_{M})
\end{align}
where $\mD = \mathtt{diag}(\mH\mathbf{1}_{M})$ is a $N \times N$ diagonal matrix 
with the \emph{hyperedge degree} $d_i = \sum_{j} H_{ij}$ of each \emph{vertex} $v_i \in V$ on the diagonal 
and $\mathbf{1}_{M} = \big[ \, 1 \; 1\; \cdots \; 1 \, \big]^{\top}$ is the vector of all ones. 

The \emph{diagonal node degree matrix} $\mD_v \in \RR^{N \times N}$ is defined as 
\begin{align}\label{eq:diag-node-degree-matrix}
\mD^v = \mathtt{diag}(\mA\,\mathbf{1}_{N}) = \mathtt{diag}((\mH\mH^{\top}-\mD_v) \mathbf{1}_{N})
\end{align}\noindent
Intuitively, $\mD = \mathtt{diag}(\mH\mathbf{1}_{M})$ is the diagonal matrix of hyperedge node degrees where $D_{ii}$ is the number of hyperedges for node $i$.
Conversely, $\mD^v = \mathtt{diag}(\mA\,\mathbf{1}_{N})$ (Eq.~\ref{eq:diag-node-degree-matrix}) is the diagonal matrix of node degrees where $D^{v}_{ii}$ is the degree of node $i$.
For instance, $D_{ii}=2$ indicates that node $i$ is in two hyperedges whereas $D^{v}_{ii}=5$ indicates that node $i$ is actually connected to five nodes among those two hyperedges. 
Hence, $D^{v}_{ii}=5$ is the size of those two hyperedges.

\subsubsection{Hyperedge Diagonal Degree Matrix}
We define the diagonal hyperedge degree matrix $\mD^{e} \in \RR^{M \times M}$ as
\begin{align}\label{eq:diag-hyperedge-degree-matrix}
\mD^{e} = \mathtt{diag}(\mH^{\top}\mathbf{1}_{N})
\end{align}
where $\mD^{e} = \mathtt{diag}(\mH^{\top}\mathbf{1}_{N}) = \mathtt{diag}(d_1^{e}, d_2^{e}, \ldots, d_M^{e})$ is a $M \times M$ diagonal matrix 
with the hyperedge degree $d^{e}_j = \sum_{i} H_{ij}$ of each hyperedge $e_j \in E$ on the diagonal 
and $\mathbf{1}_{N} = \big[ \, 1 \; 1\; \cdots \; 1 \, \big]^{\top}$.

\subsubsection{Node Adjacency Matrix}
Given $\mH$, we define the $N \times N$ node adjacency matrix $\mA$ as 
\begin{align} \label{eq:node-adj-matrix}
\mA = \mH\mH^{\top} - \mD
\end{align}
where $\mD =$ $N \times N$ vertex degree diagonal matrix with $D_{ii}=\sum_{j}H_{ij}$.

\subsubsection{Hyperedge Adjacency Matrix}
Similarly, we define the $M \times M$ hyperedge adjacency matrix $\mA^{(e)}$ as
\begin{align} \label{eq:hyperedge-adj-matrix}
\mA^{(e)} = \mH^{\top}\mH - \mD^{e}
\end{align}
where $\mD^e = $ is the $M \times M$ hyperedge degree diagonal matrix with $D^{e}_{ii}=\sum_{j}H_{ji}$.
The graph formed from Eq.~\ref{eq:hyperedge-adj-matrix} is related to the notion of an intersection graph that encodes the intersection patterns from a family of sets.
This also has an inherent connection to the \emph{line graph of a hypergraph}.
More formally, let $G_L$ denote the line graph of the hypergraph $G$ formed from the hyperedges $S_i,\; i=1,2,\ldots,M$, representing sets of vertices, 
and let $\{\delta_i\}_{i=1}^{M}$ denote the intersection thresholds for the hyperedges such that $\forall i,\, \delta_i>0$.
Then, the edge set $E_{\delta}(G_L)$ is defined as:
\begin{align} \label{eq:refined-hyperedge-edge-set-thresholded}
E_{\delta}(G_L) = \big\{\{v_i,v_j\} \,\big|\; i \not= j,\, |S_i \cap S_j| > \delta_i \big\}
\end{align}
where $v_i$ is the vertex created for each hyperedge in the hypergraph.
For simplicity, one can also set $\delta_1=\delta_2=\cdots=\delta_M$ or set $\delta_i$ to be a fixed fraction of the hyperedge size $|S_i|$.
To see the connection between the edge set in Eq.~\ref{eq:refined-hyperedge-edge-set-thresholded} to the edge set from the hyperedge adjacency matrix $\mA^{(e)}$ in Eq.~\ref{eq:hyperedge-adj-matrix}, we can rewrite the above as follows: 
\begin{align} \label{eq:refined-hyperedge-edge-set}
E(G_L) = \big\{\{v_i,v_j\} \,\big|\; i \not= j,\, S_i \cap S_j \not= \emptyset \big\}
\end{align}
Hence, the edge set in Eq.~\ref{eq:refined-hyperedge-edge-set} is equivalent to the nonzero structure (edges) of $\mA^{(e)}$ in Eq.~\ref{eq:hyperedge-adj-matrix}.
From this perspective, it is straightforward to see that Eq.~\ref{eq:refined-hyperedge-edge-set-thresholded} represents a stronger set of hyperedge interactions when $\forall i,\, \delta_i>1$ since every edge between two hyperedges are required to share at least $\delta_i$ vertices.
Hence, $|E_{\delta}(G_L)| \geq |E(G_L)|$.

\subsection{Hypergraph Features} \label{sec:approach-hypergraph-features}
The proposed framework is extremely flexible and can take as input both hyperedge and/or node features if available.
If these initial features are not available, then we can use node2vec~\cite{node2vec}, DeepGL~\cite{deepGL}, SVD~\cite{ojo2022visgnn}, among others~\cite{struc2vec,role2vec,deepwalk} for $\phi$ and $\phi_e$ discussed below.
More formally, we define the initial feature function $\phi$ as 
\begin{align}\label{eq:derive-node-feature-matrix}
\mX = \phi(\mH\mH^{\top} - \mD) \in \RR^{N \times F}
\end{align}\noindent
where $\mH$ is the hypergraph incidence matrix and $\mX$ is the low-dimensional rank-$F$ approximation of $\mH\mH^{\top} - \mD$ computed via $\phi$.
Similarly, if the initial hyperedge feature matrix $\mY$ is not given as input, then 
\begin{align}\label{eq:derive-hyperedge-feature-matrix}
\mY &= \phi(\mH^{\top}\mH - \mD^{e}) \\
&= \phi(\mA^{(e)}) 
\end{align}\noindent
Note that Eq.~\ref{eq:derive-hyperedge-feature-matrix} is only one such way to derive $\mY$, and our framework naturally supports any such technique to obtain $\mY$.
To ensure our approach is easy to use, we include the initial feature matrix inference for nodes and more importantly, hyperedges, as a component of the framework, and do not require these as input.

\subsection{Hyperedge-Dependent Convolution} \label{sec:approach-hyperedge-dependent-convolution}
Given a hypergraph $G$, our approach learns both hyperedge and node embeddings in an end-to-end fashion.
Before introducing the hyperedge-dependent convolutions, we first define the random walk transition matrices of the nodes and hyperedges.
More formally, we define $\mP \in \RR^{N \times N}$ and $\mP_e \in \RR^{M \times M}$ as
\begin{align}\label{eq:random-walk-node-transition-matrix}
\mP & = \mH\mD_e^{-1}(\mD^{-1}\mH)^{\top} \\ 
\mP_e &= (\mD^{-1}\mH)^{\top} \mH\mD_e^{-1} \label{eq:random-walk-hyperedge-transition-matrix}
\end{align}
where $\mP$ is the \emph{random walk node transition matrix} and 
$\mP_e$ is the \emph{random walk hyperedge transition matrix}.
Now we define the node and hyperedge convolution below.
First, Eq.~\ref{eq:initial-node-embeddings} initializes the node embedding matrix $\mZ^{(1)}$ whereas Eq.~\ref{eq:initial-hyperedge-embeddings} initializes the hyperedge embeddings $\mY^{(1)}$.
Note if hyperedge features $\mY$ are given as input, then Eq.~\ref{eq:initial-hyperedge-embeddings} is replaced with $\mY^{(1)} = \mY$.
Afterwards, Eq.~\ref{eq:update-node-embeddings}-\ref{eq:update-hyperedge-embeddings} defines the hypergraph convolutional layers of our model, including the node hypergraph convolutional layer in Eq.~\ref{eq:update-node-embeddings} and the hyperedge convolutional layer in Eq.~\ref{eq:update-hyperedge-embeddings}.
More formally,
\begin{align}
\mZ^{(1)} &= \mX\qquad \text{or}\qquad \mZ^{(1)} = \phi(\mH\mH^{\top} - \mD) \label{eq:initial-node-embeddings} \\ 
\mY^{(1)} &= (\mD^{-1}\mH)^{\top}\mZ^{(1)}\qquad \text{or}\qquad \mY^{(1)}=\phi(\mH^{\top}\mH - \mD^{e}) \label{eq:initial-hyperedge-embeddings} \\ 
\mZ^{(k+1)} &= \sigma\big( (\mD^{-1}\mH\mP_e\mD_e^{-1}\mH^{\top}\mD^{-1}\mZ^{(k)} + \mD^{-1}\mH\mY^{(k)})  \mW^{(k)}\big) \label{eq:update-node-embeddings} \\
\!\!\!\!\!\!\!\!\!
\mY^{(k+1)} &= \sigma\big( (\mD_e^{-1}\mH^{\top}\mP\mD^{-1}\mH\mD_e^{-1}\mY^{(k)} + \label{eq:update-hyperedge-embeddings} \\
& \qquad (\mH\mD^{-1}_{e})^{\top}\mZ^{(k+1)})  \mW^{(k)}_{e}\big) \nonumber
\end{align}\noindent
where $\mZ^{(k+1)}$ are the updated node embeddings of the hypergraph at layer $k+1$ whereas $\mY^{(k+1)}$ are the updated hyperedge embeddings at layer $k+1$.
In the above formulation, $\sigma$ is the non-linear activation function, and for simplicity is the same for Eq.~\ref{eq:update-node-embeddings}-\ref{eq:update-hyperedge-embeddings} though the framework is flexible for using different non-linear functions for the node and hyperedge convolutional layers, that is, 
\begin{align}
\mZ^{(k+1)} &= \sigma_v\big( (\mD^{-1}\mH\mP_e\mD_e^{-1}\mH^{\top}\mD^{-1}\mZ^{(k)} + \mD^{-1}\mH\mY^{(k)})  \mW^{(k)}\big) \nonumber \\
\mY^{(k+1)} &= \sigma_e\big( (\mD_e^{-1}\mH^{\top}\mP\mD^{-1}\mH\mD_e^{-1}\mY^{(k)} + (\mH\mD^{-1}_{e})^{\top}\mZ^{(k+1)})  \mW^{(k)}_{e}\big) \nonumber
\end{align}
Furthermore, $\mW^{(k)}$ and $\mW^{(k)}_{e}$ are the learned weight matrices of the $k$th layer for nodes and hyperedges, respectively. 
Most importantly, the node embeddings at each layer are updated using the hyperedge embedding matrix $\mD^{-1}\mH\mY^{(k)}$, and similarly, the hyperedge embeddings at each layer are also updated using the $(\mH\mD^{-1}_{e})^{\top}\mZ^{(k+1)}$ node embedding matrix.
The process repeats until convergence.
Discussion of other HNN model variants from the proposed framework are provided later in Section~\ref{sec:approach-variants}.

\subsubsection{Multiple Hyperedge-Dependent Embeddings}
An important advantage of our approach is that it gives rise to multiple embeddings per node, which are dependent on the hyperedges for each of the nodes in the hypergraph.
In other words, each node $i \in V$ in the hypergraph $G$ can have multiple embeddings, that is, a set of embeddings $S_i=\{\ldots,\vz_{i}^{e},\ldots\}$ where $|S_i|=d_i^e$.
Intuitively, the number embeddings of a node $i$ is equal to the number of hyperedges of that node $i$ in the hypergraph $G$.
For a node $i$ and hyperedge $e \in E$, we have the \emph{hyperedge-dependent} node embedding $\vz_{i}^{e}$ derived as
\begin{align}\label{eq:hyperedge-dependent-function}
\vz_{i}^{e} = \psi\big(\vz_{i},\, \vy_{e}\big) \in \RR^{d}
\end{align}
where $\vz_{i}$ is the node embedding of $i$, $\vy_{e}$ is the hyperedge embedding of $e$, and $\psi$ is a function computed over the concatenation of these to obtain the hyperedge dependent embedding $\vz_{i}^{e}$ of node $i$ in the hypergraph.
The above is a general formulation of this process that derives a new hyperedge-dependent embedding using a function $\psi$ that maps the general node embedding $\vz_i$ for node $i$ and the embedding of the hyperedge $\vy_e$ for hyperedge $e \in E$ in the hypergraph to a hyperedge-dependent $d$-dimensional embedding for node $i$.
Alternatively, $\psi$ can leverage a concatenated vector $[\vz_{i}\; \vy_{e}]$ as input to derive a new hyperedge-dependent embedding for node $i$. 
More formally, 
$\vz_{i}^{e} = \psi\big([\vz_{i}\; \vy_{e}]\big) \in \RR^{d}$
This is useful and provides additional flexibility since both $\vz_{i}$ and $\vy_{e}$ can be of different dimensions.
Hence, suppose node $i$ is in hyperedge $e_j$ and $e_k$, then 
\begin{align}\label{eq:hyperedge-j-dep-emb-node-i}
\vz_{i}^{e_j} = \psi([\vz_{i}\; \vy_{e_j}])\\
\vz_{i}^{e_k} = \psi([\vz_{i}\; \vy_{e_k}]) \label{eq:hyperedge-k-dep-emb-node-i}
\end{align}
From Eq.~\ref{eq:hyperedge-j-dep-emb-node-i}-\ref{eq:hyperedge-k-dep-emb-node-i}, it is straightforward to see that $\vz_{i}^{e_j}$ and $\vz_{i}^{e_j}$ are embeddings for node $i$ that fundamentally depend on the corresponding hyperedge $e_i$ and $e_j$ that node $i$ participates. 
Hence, they represent hyperedge-dependent embeddings of node $i$.
In Figure~\ref{fig:hyperedge-dependent-embedding}, we provide an intuitive overview of the proposed hyperedge-dependent embeddings and the process for deriving them.
Hence, given a node $i \in V$ that participates in $k$ different hyperedges (that is, $k=d_i^e$), then we have the following hyperedge-dependent embedding matrix 
$\mZ_i = \left[\; \vz_{i}^{e_1} \; \cdots\vspace{1mm} \vz_{i}^{e_k} \;\right] \in \RR^{k \times d}$
for node $i \in V$ in the hypergraph.
This result is important as it shows that our approach is a more powerful generalization of 
Chitra et al.~\cite{chitra2019random} that proved that for hypergraphs to be used effectively, there must be a hyperedge-dependent weight.
Since our approach is a natural generalization of that result to $d$-dimensional hyperedge-dependent weights instead of just a single weight, HNN is able to learn from such higher-order patterns present in the hypergraph.

\subsection{Model Variants}\label{sec:approach-variants}
We now introduce a few HNN model variants that we investigate later in Section~\ref{sec:exp}.

\subsubsection{HNN-$\mP^{2}$ (2-Hops)} \label{sec:HNN-variant-P2}
Let $\mP=\mH\mD_e^{-1}(\mD^{-1}\mH)^{\top} \in \RR^{N \times N}$ be the random walk transition matrix of the nodes in our hypergraph and 
$\mP_e = (\mD^{-1}\mH)^{\top} \mH\mD_e^{-1} \in \RR^{M \times M}$ is the random walk transition matrix of the hyperedges.
Then, the two-hop HNN variant is:
\begin{align}\label{eq:HNN-two-hop-variant-node}
\mZ^{(k+1)} &= \sigma_v\big( (\mD^{-1}\mH\mD_e^{-1}\mH^{\top}\mD^{-1} + \mP + \mP\mP)\, \mZ^{(k)} \mW^{(k)}\big)  \\
\mY^{(k+1)} &= \sigma_e\big( (\mD_e^{-1}\mH^{\top}\mD^{-1}\mH\mD_e^{-1} + \mP_e + \mP_e\mP_e)\, \mY^{(k)} \mW^{(k)}_{e}\big) \label{eq:HNN-two-hop-variant-hyperedge}
\end{align}
where $\mP$ ($\mP_e$) captures the 1-hop probabilities and $\mP\mP$ ($\mP_e\mP_e$) captures the 2-hop probabilities of the nodes (and hyperedges).

\subsubsection{HNN++}\label{sec:HNN-variant-plus}
This variant also leverages $\mP=\mH\mD_e^{-1}(\mD^{-1}\mH)^{\top} \in \RR^{N \times N}$ and $\mP_e = (\mD^{-1}\mH)^{\top} \mH\mD_e^{-1} \in \RR^{M \times M}$, though used in a fundamentally different fashion.
More formally,
\begin{align}\label{eq:HNN-plus-variant-node}
\mZ^{(k+1)} &= \sigma_v\big( (\mD^{-1}\mH\mP_e\mD_e^{-1}\mH^{\top}\mD^{-1} + \mP )\, \mZ^{(k)} \mW^{(k)}\big)  \\
\mY^{(k+1)} &= \sigma_e\big( (\mD_e^{-1}\mH^{\top}\mP\mD^{-1}\mH\mD_e^{-1} + \mP_e )\, \mY^{(k)} \mW^{(k)}_{e}\big) \label{eq:HNN-plus-variant-hyperedge}
\end{align}
where both $\mP$ and $\mP_e$ are used to update the node embeddings in Eq.~\ref{eq:HNN-plus-variant-node} as well as to update the hyperedge embeddings in Eq.~\ref{eq:HNN-plus-variant-hyperedge}.
To update the node embedding matrix $\mZ$, the hyperedge random walk matrix $\mP_e$ is used to weight the interactions and node embeddings used in the aggregation and updating of the individual node embeddings.
Similarly, to update the hyperedge embeddings $\mY$, the node random walk matrix $\mP$ is used to weight the node embeddings during aggregation when updating them.

\subsubsection{HNN-Wt} \label{sec:HNN-variant-def-v1}
Now we investigate a variant that leverages $\mP_e$ as $\mH\mP_e\mH^{\top} \in \RR^{N \times N}$ to update node embeddings $\mZ^{(k+1)}$ of the hypergraph, and similarly, we use $\mP$ as $\mH^{\top}\mP\mH \in \RR^{M \times M}$ to update hyperedge embeddings $\mY^{(k+1)}$.
More formally, 
\begin{align}\label{eq:HNN-def-v1-variant-node}
\mZ^{(k+1)} &= \sigma_v\big( \mD^{-1}\mH\mP_e\mH^{\top}\mD^{-1}  \mZ^{(k)} \mW^{(k)}\big)  \\
\mY^{(k+1)} &= \sigma_e\big( \mD_e^{-1}\mH^{\top}\mP\mH\mD_e^{-1} \mY^{(k)} \mW^{(k)}_{e}\big) \label{eq:HNN-def-v1-variant-hyperedge}
\end{align}
Further, we also explored using
$\mZ^{(k+1)} = \sigma_v\big( \mH\mP_e\mH^{\top} \mZ^{(k)} \mW^{(k)}\big)$ and 
$\sigma_e\big( \mH^{\top}\mP\mH \mY^{(k)} \mW^{(k)}_{e}\big)$ to update the 
node $\mZ^{(k+1)}$ and hyperedge embeddings $\mY^{(k+1)}$, respectively, 
and observed similar results.

\subsubsection{HNN-$\mH^{2}$} \label{sec:HNN-variant-H2}
This variant uses 
the weighted node adjacency matrix of the hypergraph $\mH\mH^{\top}$ combined with the \emph{random walk node transition matrix} $\mP$ of the hypergraph to obtain $\mH\mH^{\top} + \mP$, which is then used to update the node embeddings $\mZ^{(k+1)}$ 
in Eq.~\ref{eq:HNN-IncidSq-variant-node}.
Similarly, the weighted hyperedge adjacency matrix $\mH^{\top}\mH$ combined with the \emph{random walk hyperedge transition matrix} $\mP_e$ is used to update the hyperedge embeddings $\mY^{(k+1)}$ 
in Eq.~\ref{eq:HNN-IncidSq-variant-hyperedge}.
More formally,
\begin{align}\label{eq:HNN-IncidSq-variant-node}
\mZ^{(k+1)} &= \sigma_v\big( (\mH\mH^{\top} + \mP )\, \mZ^{(k)} \mW^{(k)}\big)  \\
\mY^{(k+1)} &= \sigma_e\big( (\mH^{\top}\mH + \mP_e )\, \mY^{(k)} \mW^{(k)}_{e}\big) \label{eq:HNN-IncidSq-variant-hyperedge}
\end{align}
Other slight variations of the above would be to remove the additional $\mP$ and $\mP_e$ terms in Eq.~\ref{eq:HNN-IncidSq-variant-node}-\ref{eq:HNN-IncidSq-variant-hyperedge}, or using only $\mP=\mH\mD_e^{-1}(\mD^{-1}\mH)^{\top}$ and $\mP_e = (\mD^{-1}\mH)^{\top} \mH\mD_e^{-1}$, among others.

\subsubsection{Other Variants} 
\label{sec:HNN-variant-tanh}
We also investigated other HNN variants that use the above formulations but different non-linear functions $\sigma_v$ and $\sigma_e$ for the nodes and hyperedges; \eg, 
a HNN model with tanh and other non-linear functions are investigated in Section~\ref{sec:exp}.

\begin{table*}[t!]
\centering
\vspace{-3mm}
\caption{
Hyperedge Prediction Results (AUC).
Best result for each setting is in bold.
}
\label{tab:hyperedge-prediction-results}
\vspace{-4mm}
\begin{tabular}{l c c c c c}
\toprule
\multicolumn{1}{l}{\textbf{}}   & \multicolumn{1}{c}{\textbf{Citeseer}} & \multicolumn{1}{c}{\textbf{Cora-CC}} & \multicolumn{1}{c}{\textbf{PubMed}} & \multicolumn{1}{c}{\textbf{DBLP}} &
\multicolumn{1}{c}{\textbf{Cora-CA}} \\
\midrule

\textit{GCN} & 
0.905 $\pm$ 0.02 & 
0.879 $\pm$ 0.03 & 
0.805 $\pm$ 0.02 &  
0.926 $\pm$ 0.02 & 
0.887 $\pm$ 0.04 
\\

\textit{GraphSAGE} & 
0.917 $\pm$ 0.02 & 
0.875 $\pm$ 0.02 & 
0.823 $\pm$ 0.03 & 
0.921 $\pm$ 0.01 & 
0.874 $\pm$ 0.05 
\\

\textit{HyperGCN-Fast} & 
0.869 $\pm$ 0.02 & 
0.839 $\pm$ 0.01 & 
0.827 $\pm$ 0.01 & 
0.873 $\pm$ 0.00 & 
0.783 $\pm$ 0.02 
\\

\textit{HyperGCN} & 
0.901 $\pm$ 0.02 & 
0.907 $\pm$ 0.02 & 
0.855 $\pm$ 0.03 & 
0.934 $\pm$ 0.01 & 
0.823 $\pm$ 0.06 
\\

\textit{HGNN} & 
0.841 $\pm$ 0.02 & 
0.842 $\pm$ 0.01 &  
0.826 $\pm$ 0.00 & 
0.934 $\pm$ 0.00 & 
0.903 $\pm$ 0.03 
\\

\midrule

\textit{HNN} &
\textbf{0.965 $\pm$ 0.01} & 
\textbf{0.934 $\pm$ 0.01} & 
\textbf{0.870 $\pm$ 0.00} & 
\textbf{0.979 $\pm$ 0.00} & 
\textbf{0.937 $\pm$ 0.02} 
\\

\bottomrule
\end{tabular}
\end{table*}

\subsection{Training}\label{sec:approach-training}

\subsubsection{Hyperedge Prediction} \label{sec:training-hyperedge-pred}
We now introduce the training objective for hyperedge prediction.
Let $E=\{e_1,e_2,\ldots\}$ denote the set of known hyperedges in the hypergraph $G$ where every hyperedge $e_t = \{s_1,\ldots,s_k\} \in E$ represents a set of nodes that can be of any arbitrary size $k=|e_t|$.
Hence, for any two hyperedges $e_t, e_t^{\prime} \in E$, then $|e_t|\not=|e_t^{\prime}|$ may hold.
Further, let $F$ be a set of sampled vertex sets from the set $2^V - E$ of unknown hyperedges.
Given an arbitrary hyperedge $e \in E \cup F$, we define a hyperedge score function $f$ as:
\begin{align}
f :\;\; e=\{\vx_1,\ldots,\vx_k\} \; \to \; w
\end{align}
Hence, $f$ is a hyperedge score function that maps the set of $d$-dimensional node embedding vectors $\{\vx_1,\ldots,\vx_k\}$ of the hyperedge $e$ to a score $f(e=\{\vx_1,\ldots,\vx_k\})$ or simply $f(e)$.
Notably, HNN is flexible for use with a wide range of hyperedge score functions.
We discuss a few important hyperedge score functions later in this section; see Eq.~\ref{eq:hyperedge-score-func} and Eq.~\ref{eq:diff-max-min-eval-metric} for a few such possibilities.
Then, the hyperedge prediction loss function is:
\begin{align}
\mathbcal{L} = -\frac{1}{|E \cup F|} \sum_{e \in E \cup F} 
Y_e \log\big(\rho(f(e_t))\big) +
(1-Y_e) \log\big(1-\rho(f(e_t))\big)
\nonumber
\end{align}
where $Y_e=1$ if $e \in E$ and otherwise $Y_e=0$ if $e \in F$.
Further, we define $\rho(f(e_t))$ as
\begin{align} \label{eq:prob-hyperedge-t-sigmoid}
\rho(f(e_t)) = \frac{1}{1+\exp[-f(e_t)]}
\end{align}
where $p(e_t) = \rho(f(e_t))$ is the probability of hyperedge $e_t$ existing in the hypergraph $G$.

Now we formally define a few useful hyperedge score functions $f$ that we investigate in this work.
The hyperedge score $f(e)$ can be derived as the mean cosine similarity between any pair of nodes in the hyperedge $e \in E$ as follows:
\begin{align}\label{eq:hyperedge-score-func}
f(e) = \frac{1}{T} \sum_{\substack{i,j \in e \\ \text{s.t. } i>j}} \vx_i^{\top}\vx_j
\end{align}
where $T=\frac{|e|(|e|-1)}{2}$ is the number of unique node pairs $i,j$ in the hyperedge $e$.
Intuitively, the hyperedge score $f(e)$ is largest when all nodes in the set $e=\{s_1,s_2,\ldots\}$ have similar embeddings.
In the extreme, $f(e)\rightarrow 1$ implies $\vx_i^{\top}\vx_j=1$ for all $i,j \in e$.
Conversely, when $f(e)\rightarrow 0$, then $\vx_i^{\top}\vx_j=0$ for all $i,j \in e$, implying that the set of nodes in the hyperedge is independent with orthogonal embedding vectors.
When $0<f(e)<1$ lies between these two extremes, this indicates intermediate similarity or dissimilarity.

Alternatively, we may also define a hyperedge score function $f$ based on the difference between the max and min value over the set of nodes in the hyperedge $e$.
More formally, suppose we have a hyperedge $e$ with $k$ nodes, $\vx_1,\ldots,\vx_k \in \RR^{d}$, then
\begin{align}\label{eq:diff-max-min-eval-metric}
f(e) = \max_{i \in [k]} \vx_i - \min_{j \in [k]} \vx_j
\end{align}
where $f(e)$ is the difference between its maximum and minimum value over all nodes in the hyperedge $e$.

\subsubsection{Node Classification} \label{sec:training-node-classification}
Given a hypergraph $G=(V,E)$ along with a small set of labeled nodes $V_L$, the goal of semi-supervised node classification is to predict the remaining labels of the nodes $V \setminus V_L$.
Then, the hypergraph node classification loss $\mathbcal{L}$ is:
\begin{align}\label{eq:node-classification}
\mathbcal{L} = - \frac{1}{|V_L|} \sum_{i \in V_L} 
\sum_{k=1}^{|C|} Y_{ik} \log P_{ik}
\end{align}
where $Y_{ik}$ corresponds to the $k$-th element of the one-hot encoded label for node $i \in V_L$, that is, $\vy_i \in \{0,1\}^{|C|}$ and $P_{ik}$ is the predicted probability of node $i$ being labeled class $k$.

\begin{table}[h!]
\centering
\vspace{-2mm}
\caption{
Hyperedge prediction results for a variety of HNN variants from the proposed framework.
}
\label{tab:hyperedge-prediction-results-variants}
\vspace{-4mm}
\begin{tabular}{@{}l@{} @{}c@{} c@{} c@{} c@{} c@{}@{}}
\toprule
& 
\multicolumn{1}{c}{\textbf{\!\!\!\!\!Citeseer}} &
\multicolumn{1}{c}{\textbf{\!\!\!Cora-CC}} &
\multicolumn{1}{c}{\textbf{\!\!PubMed\!\!}} &
\multicolumn{1}{c}{\textbf{\!\!DBLP\!}} &
\multicolumn{1}{c}{\textbf{\!\!Cora-CA\!\!}} \\
\midrule

\textit{HNN} &
0.965  & 
0.934  & 
0.870  & 
0.979  & 
\textbf{0.937}  
\\

\textit{HNN-$\mP^2$} (\S\ref{sec:HNN-variant-P2}) &
0.966  & 
0.934  &
0.898  & 
\textbf{0.981}  & 
0.932  
\\

\textit{HNN++} \;(\S\ref{sec:HNN-variant-plus}) &
\textbf{0.967}  &
0.917  & 
0.879  & 
0.965  & 
0.923  
\\

\textit{HNN-Wt} (\S\ref{sec:HNN-variant-def-v1}) &
0.954  & 
0.890  & 
0.869  & 
0.971  & 
0.907  
\\

\textit{HNN-$\mH^{2}$} (\S\ref{sec:HNN-variant-H2}) &
\textbf{0.967}  & 
\textbf{0.936}  & 
0.897  & 
\textbf{0.981}  &
0.925  
\\

\textit{HNN-$\tanh$} (\S\ref{sec:HNN-variant-tanh}) &
0.959  & 
0.926  &
\textbf{0.918}  & 
0.925 & 
0.913  
\\

\bottomrule
\end{tabular}
\vspace{-3mm}
\end{table}

\section{Experiments}\label{sec:exp}
The experiments are designed to investigate the research questions:
\begin{itemize}[leftmargin=*]
\item \textbf{RQ1:} Does HNN achieve better performance for hyperedge prediction compared to 
the other models (Section~\ref{sec:exp-hyperedge-pred-results})?

\item \textbf{RQ2:} Can HNN achieve better performance over other models for node classification on hypergraphs (Section~\ref{sec:node-classification-results})?

\item \textbf{RQ3:} Does HNN achieve better data efficiency compared to the other models (Section~\ref{sec:exp-data-efficiency})? 

\item \textbf{RQ4:} How do the hyperparameters and different components of HNN impact the performance
(Section~\ref{sec:ablation-study})?
\end{itemize}
\noindent
For the experiments, we use the network data
from Yadati et al.~\cite{hypergcn_neurips19}.\unskip\footnote{https://github.com/malllabiisc/HyperGCN} 
We summarize the datasets and their properties in Table~\ref{table:dataset-stats-public}.

\subsection{Hyperedge Prediction Results}\label{sec:exp-hyperedge-pred-results}
To answer RQ1, we investigate the effectiveness of HNN for hyperedge prediction.
For training, we select $p\%$ of the observed hyperedges and use the remaining $1-p\%$ for testing. Unless otherwise mentioned, we use $80\%$ of the hyperedges for training and $20\%$ for testing. 
Now, we sample the same number of negative hyperedges as follows: we uniformly select an observed hyperedge $e \in E$, and derive a corresponding negative hyperedge $f \in F$ by sampling uniformly at random $\frac{|e|}{2}$ nodes from $e \in E$ and sampling the other $\frac{|e|}{2}$ nodes from $V - e$. 
This generates negative hyperedges that are more challenging to differentiate compared to uniformly sampling a set of $|e|$ nodes from $V$. 
We compared HNN against 
HyperGCN and HyperGCN-fast~\cite{hypergcn_neurips19}, HGNN~\cite{HGNN}, GraphSAGE~\cite{ying2018graph}, GCN~\cite{kipf2016semi}, and a simple MLP.
For all methods, we used the author suggested hyperparameters and their implementations.
We report mean AUC and standard deviation for each dataset over 10 trials.
The results are provided in Table~\ref{tab:hyperedge-prediction-results}. 
Overall, HNN achieves the best performance over all other models and across all datasets investigated as shown in Table~\ref{tab:hyperedge-prediction-results}.
HNN achieves an overall mean gain of 7.72\% across all models and graphs.
Notably, HNN achieves a relative mean gain 
of 
$11.88\%$ over HyperGCN-Fast, 
$7.92\%$ over HGNN, 
$6.46\%$ over GCN, 
$6.24\%$ over GraphSAGE, and 
$6.10\%$ over HyperGCN.
These results demonstrate the effectiveness of HNN for hyperedge prediction.

\begin{table*}[t!]
\centering
\vspace{-2.3mm}
\caption{
Node Classification Results.
Best result for each setting is in bold.
}
\label{table:node-classification-results}
\vspace{-4mm}
\begin{tabular}{l ccccc}
\toprule
\multicolumn{1}{l}{\textbf{}}   & \multicolumn{1}{c}{\textbf{Citeseer}} & \multicolumn{1}{c}{\textbf{Cora-CC}} & \multicolumn{1}{c}{\textbf{PubMed}} & \multicolumn{1}{c}{\textbf{DBLP}} &
\multicolumn{1}{c}{\textbf{Cora-CA}} 
\\
\midrule

\textit{MLP} & 
0.746 $\pm$ 0.01 & 
0.755 $\pm$ 0.01 & 
0.803 $\pm$ 0.01 & 
0.905 $\pm$ 0.01 & 
0.755 $\pm$ 0.01 \\

\textit{GCN} & 
0.775 $\pm$ 0.01 &
0.764 $\pm$ 0.01 & 
0.802 $\pm$ 0.01 & 
0.921 $\pm$ 0.02 & 
0.813 $\pm$ 0.01  \\

\textit{GraphSAGE} & 
0.759 $\pm$ 0.01 &
0.801 $\pm$ 0.02 & 
0.819 $\pm$ 0.02 & 
0.851 $\pm$ 0.02 & 
0.811 $\pm$ 0.01  \\

\textit{HyperGCN-Fast} &
0.802 $\pm$ 0.04 & 
0.775 $\pm$ 0.05 & 
0.805 $\pm$ 0.05 & 
0.953 $\pm$ 0.02 & 
0.841 $\pm$ 0.05 
\\

\textit{HyperGCN} &
0.849 $\pm$ 0.01 & 
0.734 $\pm$ 0.04 & 
0.820 $\pm$ 0.06 & 
0.947 $\pm$ 0.01 & 
0.711 $\pm$ 0.08 \\    

\textit{HGNN} &
0.651 $\pm$ 0.04 & 
0.700 $\pm$ 0.03 & 
0.538 $\pm$ 0.03 & 
0.965 $\pm$ 0.02 & 
0.837 $\pm$ 0.05 
\\

\midrule

\textit{HNN} & 
\textbf{0.877 $\pm$ 0.01} & 
\textbf{0.803 $\pm$ 0.02} &
\textbf{0.841 $\pm$ 0.01} & 
\textbf{0.981 $\pm$ 0.00} & 
\textbf{0.916 $\pm$ 0.00} 
\\

\bottomrule
\end{tabular}
\end{table*}

\begin{table}[h!]
\centering
\vspace{-2mm}
\caption{
Node classification results for HNN variants.
}
\label{table:node-classification-variant-results}
\vspace{-4mm}
\begin{tabular}{@{}l@{} @{}c@{} c@{} c@{} c@{} c@{}@{}}
\toprule
& 
\multicolumn{1}{c}{\textbf{\!\!\!\!\!Citeseer}} & 
\multicolumn{1}{c}{\textbf{\!\!\!Cora-CC}} &
\multicolumn{1}{c}{\textbf{\!\!PubMed\!\!}} &
\multicolumn{1}{c}{\textbf{\!\!DBLP\!}} &
\multicolumn{1}{c}{\textbf{\!\!Cora-CA\!\!}}  \\
\midrule

\textit{HNN}					& 
\textbf{0.877}  & 
0.803  & 
\textbf{0.841}  & 
\textbf{0.981}  & 
0.916   \\

\textit{HNN-$\mP^2$} (\S\ref{sec:HNN-variant-P2}) &
0.822  &
0.885  & 
0.835  & 
\textbf{0.981}  & 
0.913   \\

\textit{HNN++} \;(\S\ref{sec:HNN-variant-plus}) &
0.831  & 
\textbf{0.891}  &
\textbf{0.841}  & 
\textbf{0.981}  & 
0.916  \\

\textit{HNN-Wt} (\S\ref{sec:HNN-variant-def-v1}) &
0.833  & 
0.886  & 
0.840  & 
0.980  & 
\textbf{0.920}    \\

\bottomrule
\end{tabular}
\vspace{-4mm}
\end{table}

While HNN was shown in Table~\ref{tab:hyperedge-prediction-results} to achieve significantly better hyperedge predictive performance across all other methods, we now investigate other HNN variants from the proposed framework to better understand the effectiveness of various components of the framework and whether these different HNN models may lead to even better predictive performance for the hyperedge prediction task.
Results are provided in Table~\ref{tab:hyperedge-prediction-results-variants}.
Notably, we find that the HNN variants achieve even better AUC across all but one graph as shown in Table~\ref{tab:hyperedge-prediction-results-variants}.
Furthermore, the difference in performance between HNN and these variants are sometimes very large as shown in Table~\ref{tab:hyperedge-prediction-results-variants} for PubMed.
While even simple HNN variants outperform the other models (Table~\ref{tab:hyperedge-prediction-results}), there is often an HNN variant that can perform even better for the specific graph of interest.
Future work should further explore the proposed hypergraph representation learning framework and other interesting variants that arise from it.
We also demonstrated the effectiveness of HNN for different loss functions (Table~\ref{tab:hyperedge-prediction-results-vary-scoring-func}) in Appendix and as the difficulty of the hyperedge prediction task increases (Figure~\ref{fig:hyperedge-pred-diff}).

\begin{figure}[t!]
\vspace{-3mm}
\centering
\hspace{-4mm}
\subfigure{
\includegraphics[width=0.50\linewidth]{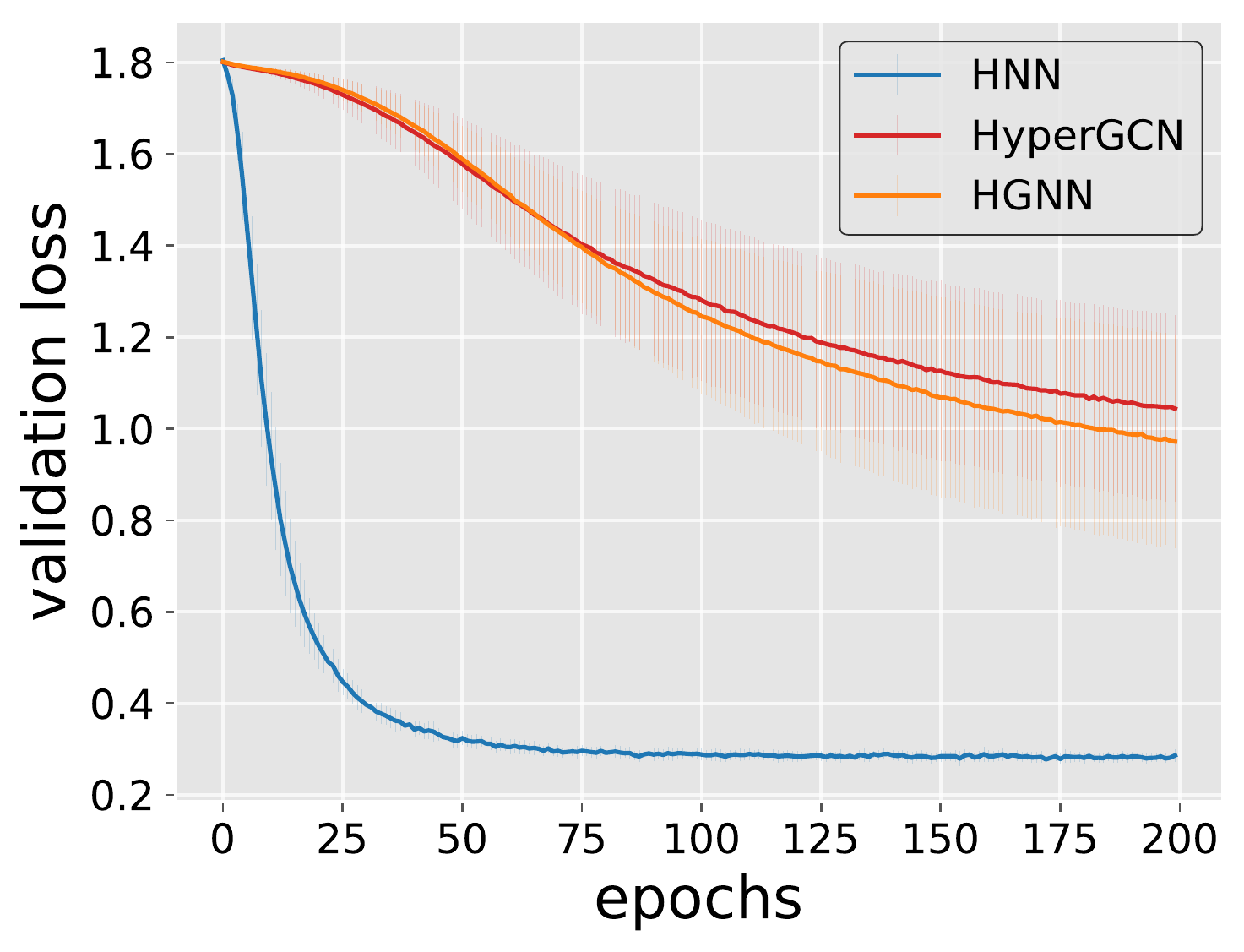}
\label{fig:loss-curve-DBLP}
}
\hspace{-4mm}
\subfigure{
\includegraphics[width=0.50\linewidth]{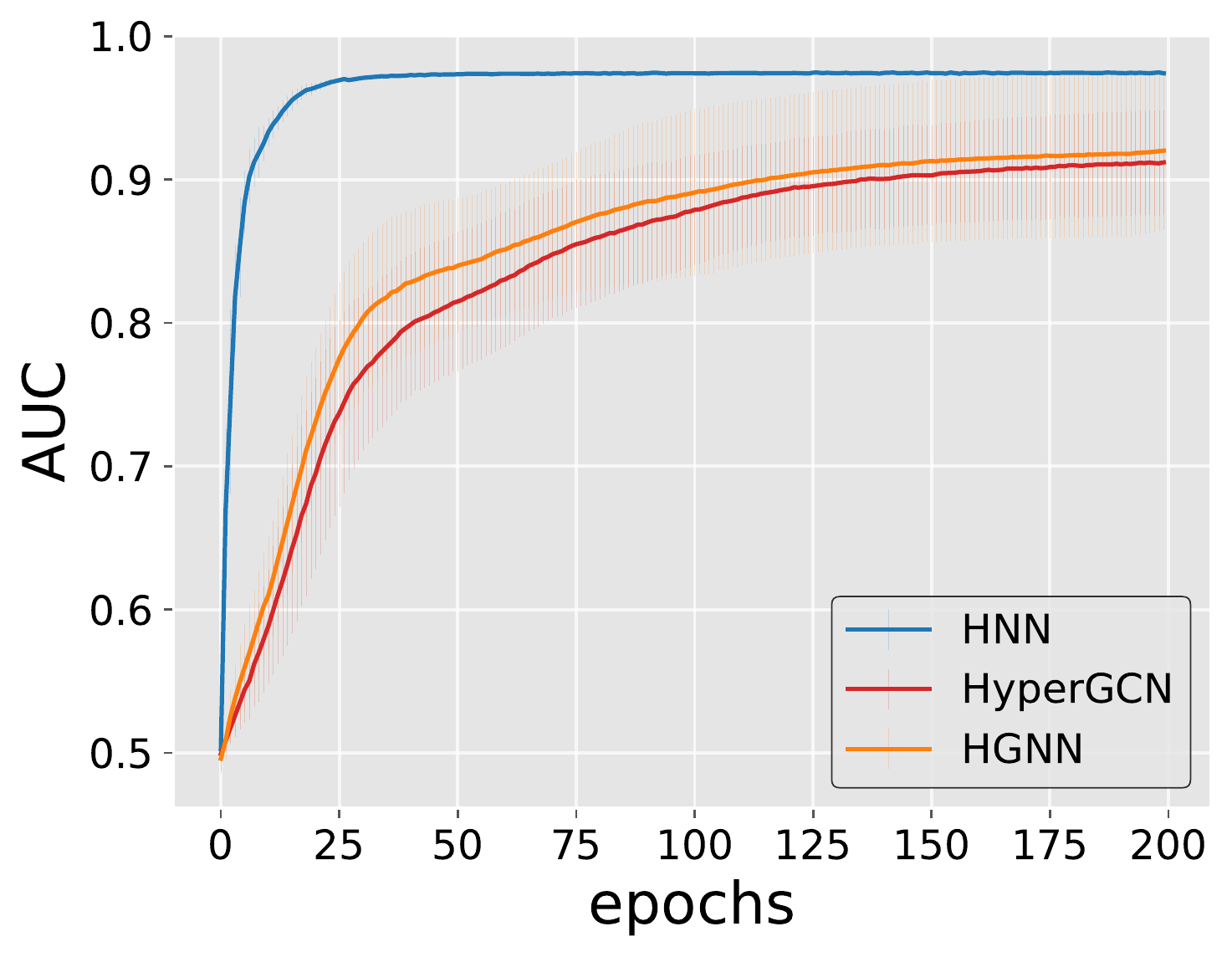}
\label{fig:AUC-curve-DBLP}
}
\vspace{-5mm}
\caption{
Data Efficiency Results. HNN achieves significantly lower loss and better AUC across all epochs compared to HyperGCN and HGNN on DBLP.
See text for discussion.
}
\label{fig:learning-curves}
\vspace{-4mm}
\end{figure}

\subsection{Node Classification Results}\label{sec:node-classification-results}
We now investigate the HNN framework for node classification in hypergraphs (RQ2).
For this task, we use the train-test splits of Yadati et al.~\cite{hypergcn_neurips19} and the same baselines as before.
All baselines are trained with default parameters. 
We use AUC for evaluation and report the mean and standard deviation for each dataset over the 10 train-test splits.
The results are provided in Table~\ref{table:node-classification-results}. 
Overall, we find that HNN achieves the best performance across all graphs as shown in Table~\ref{table:node-classification-results}. 
Notably, HNN achieves a mean gain in AUC of $11.37\%$ over all models and across all benchmark hypergraphs.
In particular, HNN achieves a mean gain of 
$23.37\%$ over HGNN, 
$11.67\%$ over MLP, 
$9.54\%$ over HyperGCN, 
$9.34\%$ over GraphSAGE, 
$8.46\%$ over GCN, and 
$5.86\%$ over HyperGCN-fast. 
Nevertheless, the results in Table~\ref{table:node-classification-results} demonstrate the effectiveness of HNN for hypergraph node classification.
We now explore a few other HNN variants in Table~\ref{table:node-classification-variant-results}.
Notably, these variants can sometimes achieve even better predictive performance compared to the simple HNN model compared previously in Table~\ref{table:node-classification-results}.
For instance, while HNN achieves an AUC of 0.803 for Cora-CC, it performs the worst compared to the HNN variants investigated as shown in Table~\ref{table:node-classification-variant-results}.
Furthermore, HNN++ achieves an AUC of 0.891 compared to 0.803 using HNN, which is a significant improvement.
Hence, the proposed HNN framework provides a powerful basis for developing even better models for hypergraph representation learning tasks.

\subsection{Data Efficiency Results}\label{sec:exp-data-efficiency}
We now explore the \emph{data efficiency} and predictive power of HNN (RQ3) in Figure~\ref{fig:learning-curves}.
Overall, we find that HNN is significantly more data efficient compared to 
HyperGCN and HGNN across all epochs.
For instance, at 25 epochs, the loss and AUC of HNN are around 0.4 and 0.96, respectively, whereas the loss and AUC of HyperGCN and HGNN are around 1.7 and at most 0.8, respectively. 
Hence, using only 25 epochs, HNN achieves around 4x better loss and around 20\% gain in AUC as shown in Figure~\ref{fig:learning-curves}.
Due to space, results for other datasets were removed, though similar findings were observed.
It is also straightforward to see that HNN has significantly lower standard error compared to HyperGCN and HGNN as shown in Figure~\ref{fig:learning-curves}.
This is another important property of efficient and consistent training algorithms.
Notably, HNN is the only model that simultaneously learns an embedding for each hyperedge as well as an embedding for each node of the hyperedge.
We posit that it is precisely this fact that enables HNN to achieve such significant improvement on both predictive power as well as data efficiency.

\subsection{Hyperparameter Sensitivity} \label{sec:ablation-study}
\begin{figure}[h!]
\vspace{-2mm}
\centering
\hfill
\hspace{-3mm}
\subfigure{
\includegraphics[width=0.50\linewidth]{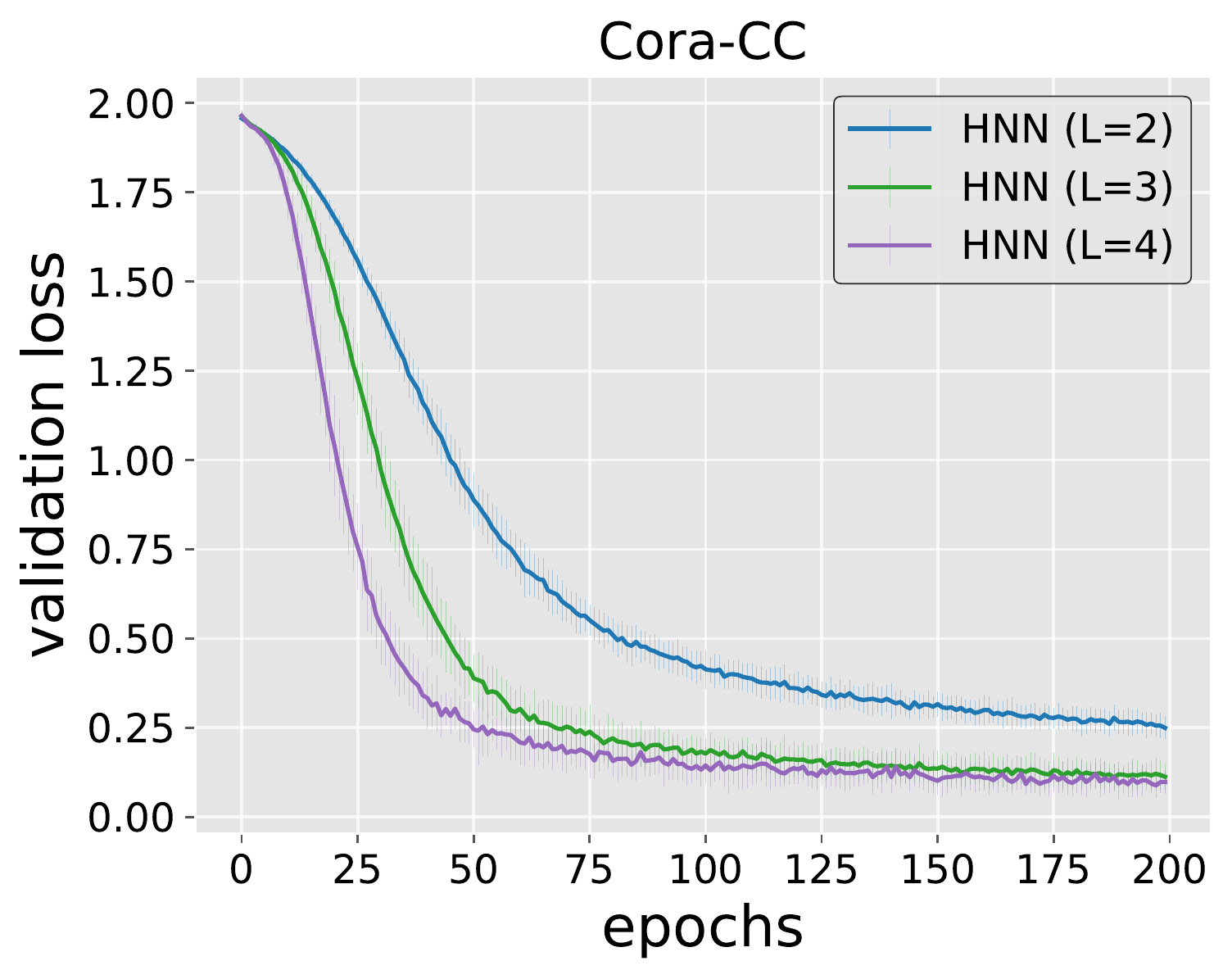}
\label{fig:loss-curve-Cora-CC-vary-layers}
}
\hfill
\hspace{-4mm}
\subfigure{
\includegraphics[width=0.50\linewidth]{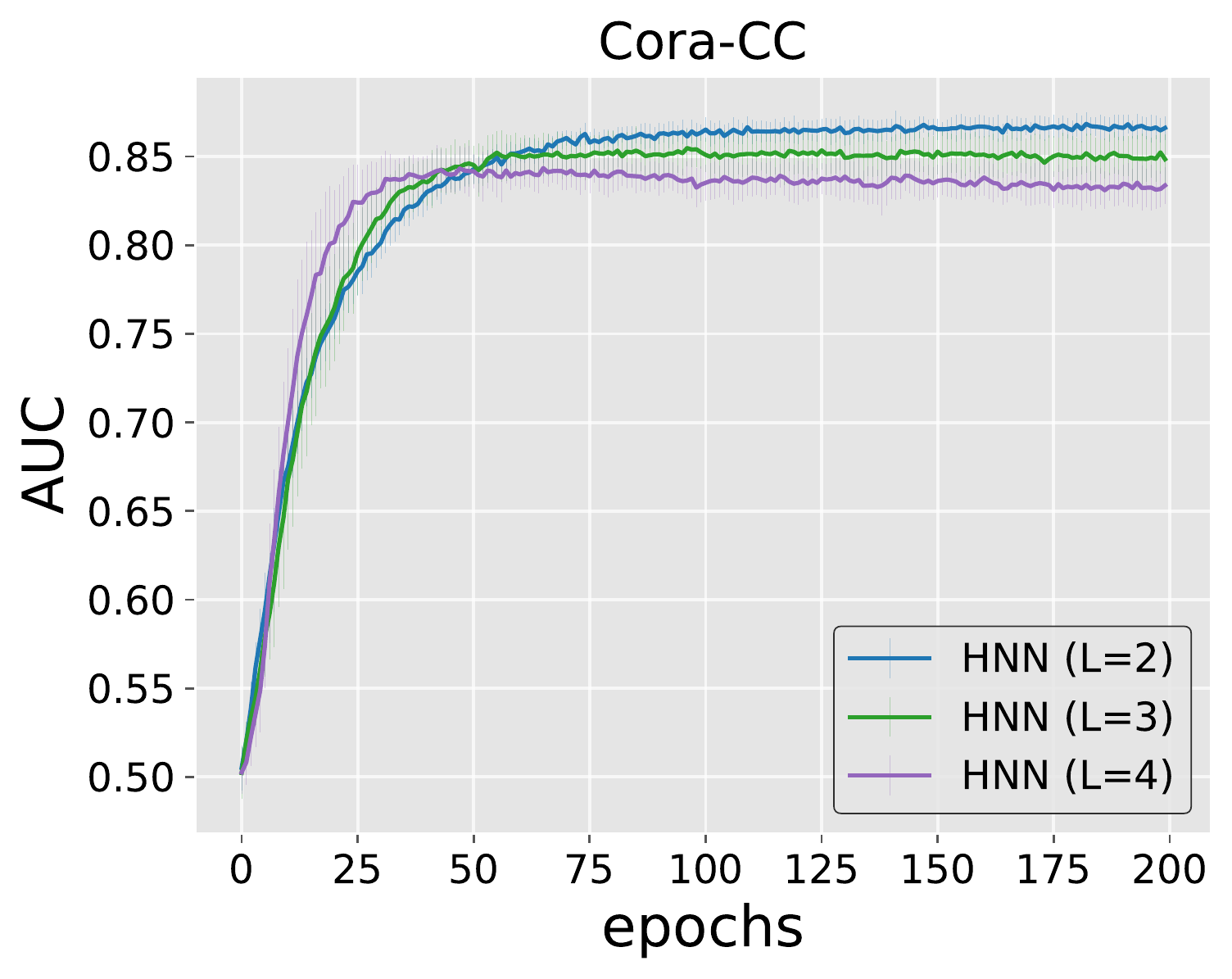}
\label{fig:AUC-curve-Cora-CC-vary-layers}
}

\vspace{-5.5mm}
\caption{
Loss and AUC learning curves as the number of layers increases from 2 to 4.
}
\label{fig:learning-curves-vary-layers-Cora-CC}
\vspace{-3mm}
\end{figure}

We first investigate the choice of non-linear function $\sigma$ on the performance of HNN in Table~\ref{table:ablation-nonlinear-func}.
From Table~\ref{table:ablation-nonlinear-func}, we see that HNN with gelu performs the best on 3 of the 5 benchmark hypergraphs, whereas tanh and RReLU perform the best on the other 2 hypergraphs.
These results indicate that HNN can achieve even better predictive performance when configured with the appropriate non-linear function.
In addition, we also investigate the loss and AUC as the number of epochs increases for HNN models with layers $L\in\{2,3,4\}$.
In Figure~\ref{fig:learning-curves-vary-layers-Cora-CC}, we observe that when only 25 epochs are used, HNN with $L=4$ layers achieves the best AUC with lower loss.
However, as the number of epochs increases, the other HNN models trained using fewer layers become competitive.
See Appendix for many other results removed due to space.

\begin{table}[h!]
\centering
\vspace{-2mm}
\caption{
Varying non-linear functions $\sigma$ of HNN.
}
\label{table:ablation-nonlinear-func}
\vspace{-4mm}
\begin{tabular}{@{}l@{} c@{} c@{} c@{} c@{} c@{}}
\toprule
\multicolumn{1}{l}{\textbf{$\sigma$}}   & 
\multicolumn{1}{c}{\textbf{\!\!\!\!\!Citeseer}} &
\multicolumn{1}{c}{\textbf{Cora-CC}} &
\multicolumn{1}{c}{\textbf{PubMed}} &
\multicolumn{1}{c}{\textbf{DBLP}} &
\multicolumn{1}{c}{\textbf{Cora-CA\!\!}} 
\\
\midrule

tanh & 
0.829  & 
0.885  & 
0.842  & 
\textbf{0.982}  &
0.916   \\

leakyReLU &
0.833  & 
0.885  & 
0.840  & 
0.980  & 
0.920  
\\

gelu &
0.840  &
\textbf{0.890}  &
\textbf{0.843}  & 
0.980  & 
\textbf{0.921}   \\

selu &
0.826  & 
0.884  & 
0.837  & 
0.980  & 
0.917  \\

RReLU &
\textbf{0.877}  &
0.803 & 
0.841  &
0.981  & 
0.916   \\

\bottomrule
\end{tabular}
\vspace{-2mm}
\end{table}

\section{Case Study: Style Recommendation}\label{sec:email-style-recommendation}
We now investigate our proposed hypergraph neural network framework for the application of design style recommendation for HTML documents, which could be websites, posters, and marketing emails. 
For this novel application, we first collected a large-scale HTML document corpus from Really Good Emails (\url{https://reallygoodemails.com/}), and then extract the HTML fragments (i.e., email sections) from each email document in the corpus (Figure~\ref{fig:hypergraph-model}).
Such fragments may consist of buttons, background-style, text, images, and so on.
We summarize the dataset and its properties in Table~\ref{table:data-stats} along with examples of the entity types extracted in Table~\ref{table:entity-type-summary} in Appendix. 
Due to the uniqueness of words and images in individual email, they are extracted  but not used in style learning.
To make it easy for others to investigate this new style recommendation task, we release the corpus of HTML emails along with the heterogeneous hypergraph derived from it:
\begin{center}
\url{https://networkrepository.com/style-rec}
\end{center}

\subsection{Hypergraph Extraction}
\begin{figure}[t!]
\centering
\vspace{-3mm}
\includegraphics[width=0.95\linewidth]{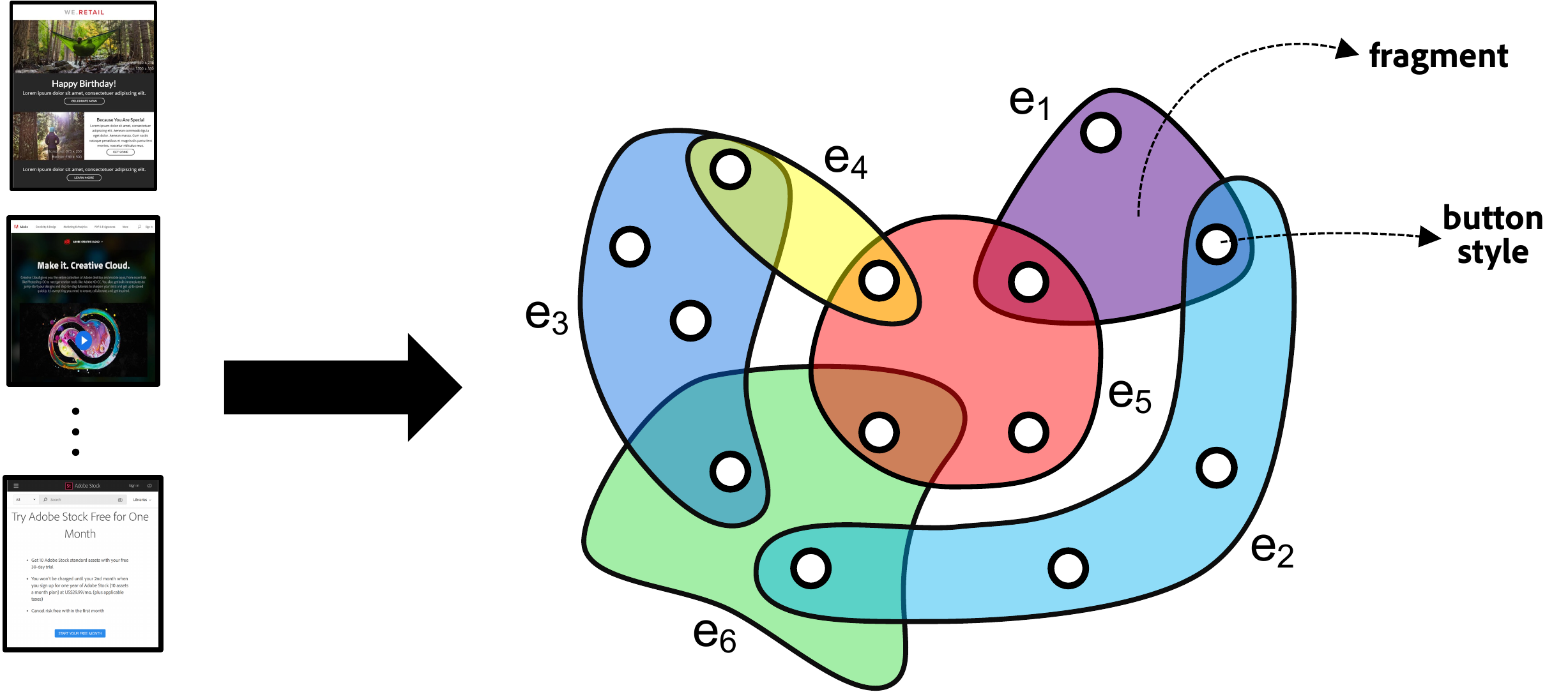}
\vspace{-3mm}
\caption{
Given the corpus of HTML email documents, we extract a large hypergraph that succinctly encodes the dependencies between the various sets of entities.
}
\label{fig:hypergraph-model}
\vspace{-3mm}
\end{figure}

Given a large corpus of HTML documents (\ie, promotional marketing emails), we extract a large heterogeneous hypergraph from the corpus by first decomposing each HTML email into a set of fragments as shown in Figure~\ref{fig:extraction-example} in Appendix.
Now, for every fragment, we decompose it further into smaller fine-grained entities such as buttons, background style, and so forth (Table~\ref{table:entity-type-summary}). 
These entities are included as nodes in the hypergraph and the set of all entities extracted from the fragment are encoded as a hyperedge as shown in Figure~\ref{fig:hypergraph-model}.
Hence, every hyperedge in the heterogeneous hypergraph represents a fragment from some HTML document.
To capture the spatial relationship present between fragments in an HTML document, we also include a node for each fragment along with an edge connecting each fragment to the fragment immediately below or beside it.
Notice that hyperedges in this hypergraph are heterogeneous in that they consist of a set of heterogeneous nodes of various types 
as shown in Table~\ref{table:data-stats}.
Most importantly, the entities of a fragment (hyperedge) are not unique to the specific fragment, and can be connected to a wide variety of other fragments (hyperedges) as shown in Figure~\ref{fig:hypergraph-model}.
Notably, we see that two HTML fragments represented as hyperedges $e_1$ and $e_2$ in Figure~\ref{fig:hypergraph-model} contain buttons with the same style.
This overlap in button-style often implies other stylistic similarities between the two fragments.

\begin{table}[h]
\vspace{-1.5mm}
\caption{
Statistics and properties of our HTML document (email) corpus and the resulting heterogeneous.
}
\label{table:data-stats}
\vspace{-4mm}
\small
\begin{tabular}{l cccc}
\multicolumn{5}{@{}p{\linewidth}@{}}{
Note $|V|$ denotes the number of nodes of a given node type;
$\Delta$ denotes the max hyperedge degree; 
$d_{\rm avg}$ and $d_{\rm med}$ are the average and median degree.
} \\ 
\toprule
\textsc{Node Type} & $|V|$  & $\Delta$ & $d_{\rm avg}$ & $d_{\rm med}$ 
\\ \midrule

\textbf{button-style} & 2361  & 21 & 7.14 & 6 \\
\textbf{text-style} & 14678  & 1548 & 173.42 & 6\\
\textbf{background \& font color} & 2798 &  72 & 10.46 & 4\\
\textbf{background-style} & 811 &  73 & 20.96 & 8\\
\textbf{word} & 31761 &  6664 & 382.77 & 22\\
\textbf{image} & 7682 &  200 & 7.29 & 1\\
\textbf{fragment} & 24614 &  235 & 27.84 & 16\\

\bottomrule
\end{tabular}
\vspace{-4.0mm}
\end{table}

\begin{table}[t!]
\caption{Results for Button Style Recommendation.}
\label{table:results-style-button}
\vspace{-4mm}
\small
\begin{tabular}{l cccc}
\toprule
&
\multicolumn{4}{c}{HR@K} \\ 
\cmidrule(lr){2-5} 
\textbf{Model} & 
@1 & @10 & @25 & @50 
\\ \midrule
Random & 0.000 $\pm$ 0.00 & 0.000 $\pm$ 0.00 & 0.018 $\pm$ 0.00& 0.027 $\pm$ 0.00 \\ 
Pop. & 0.000 $\pm$ 0.00 & 0.000 $\pm$ 0.00  & 0.009 $\pm$ 0.00& 0.009 $\pm$ 0.00 \\ 

HyperGCN 
& 0.008 $\pm$ 0.01
& 0.011 $\pm$ 0.00
& 0.043 $\pm$ 0.02
& 0.066 $\pm$ 0.03
\\ 
\midrule

HNN 
& 0.243 $\pm$ 0.05
& 0.477 $\pm$ 0.03
& 0.536 $\pm$ 0.06
& 0.594 $\pm$ 0.05 
\\ 
\bottomrule
\end{tabular}
\vspace{-0mm}
\end{table}

\begin{table}[t!]
\vspace{-1.5mm}
\caption{Results for Background Style Recommendation.}
\label{table:results-style-background}
\vspace{-4mm}
\small
\begin{tabular}{l cccc}
\toprule
&
\multicolumn{4}{c}{HR@K} 
\\ 
\cmidrule(lr){2-5} 
\textbf{Model} & @1  & @10 & @25 & @50 
\\ \midrule
Random & 0.000 $\pm$ 0.00&  0.000 $\pm$ 0.00& 0.000 $\pm$ 0.00& 0.074 $\pm$ 0.00\\ 

Pop. & 0.000 $\pm$ 0.00 & 0.000 $\pm$ 0.00& 0.000 $\pm$ 0.00& 0.000 $\pm$ 0.00\\ 

HyperGCN & 0.000 $\pm$ 0.00 & 0.031 $\pm$ 0.04& 0.061 $\pm$ 0.06& 0.147 $\pm$ 0.14\\ 

\midrule

HNN & 0.181 $\pm$ 0.11 & 0.457 $\pm$ 0.14& 0.552 $\pm$ 0.10& 0.741 $\pm$ 0.08\\ 

\bottomrule
\end{tabular}
\vspace{-2.8mm}
\end{table}

\subsection{Quantitative Recommendation Results} 
To quantitatively evaluate the effectiveness of our approach for style recommendation tasks, we hold out $20\%$ of links in the hypergraph that occur between a fragment and a specific style entity (e.g., button-style) to use as ground-truth for quantitative evaluation.
In particular, suppose we are leveraging our approach for recommending button-styles, then we uniformly at random select $20\%$ of the links that occur between a fragment and a button-style for testing.
Then HNN is trained using the training graph which does not contain the $20\%$ of held-out links.
Now, we derive a score between fragment $i$ and every button-style $k \in V_B$ using the learned embeddings from our approach:
\begin{align}
\vw_i = f(\vz_i, \vz_k), \quad \forall k \in V_B
\end{align}
where $f$ is a score function (\ie, cosine) and $\vw_i=\big[\vw_{i1} \; \cdots \; \vw_{i|V_B|}\,\big]$ are the scores.
We then sort the scores $\vw_i$ and recommend top-K styles with largest weight.
To quantitatively evaluate the performance of our approach for this ranking task, we use HR@K and nDCG@K where $K=\{1,10,25,50\}$.
We repeat the above for each of the held-out links in the test set (e.g., between a fragment and button-style) and report the average of the evaluation metrics. 
Since this is the first work to study this problem, there are no immediate methods for comparison.
Nevertheless, we use common-sense baselines including
random that recommends a design style element uniformly at random among the set of possibilities,
popularity (pop) that recommends the most frequent style,
and HyperGCN.

The results showing the effectiveness of the various approaches for recommending the top button-styles are provided in Table~\ref{table:results-style-button}.
Due to space constraints, nDCG results are provided in Table~\ref{table:results-style-button-nDCG} in Appendix.
Notably, HNN performs significantly better than the other models across both HR@K and nDCG@K for all $K\in\{1,10,25,50\}$.
In many instances, the simple random and popularity baseline are completely ineffective with HR@K and nDCG@K of 0 when $K$ is small (top-1 or 10).
In contrast, HNN is able to recover the ground-truth button-style 24\% of the time in the top-1 as shown in Table~\ref{table:results-style-button}.
Now we investigate HNN for recommending useful background-styles.
Results for HR@K are reported in Table~\ref{table:results-style-background}; see Table~\ref{table:results-style-background-nDCG} in Appendix for nDCG results.
Our approach achieves a significantly better HR and nDCG across all $K$ as shown in Table~\ref{table:results-style-background}.
It is important to note that results at smaller $K$ are more important, and these are precisely the situations where the other models completely fail.

\section{Conclusion} \label{sec:conc}
This work proposed a hypergraph representation learning framework called HNN that simultaneously learns hyperedge embeddings along with a set of hyperedge-dependent embeddings for each node in the underlying hypergraph. Notably, HNN is carefully designed to be flexible with many interchangeable components, representationally powerful for learning a set of hyperedge-dependent embeddings for each node in the hypergraph, data-efficient, and accurate with state-of-the-art performance for a wide range of hypergraph learning tasks. The experimental results demonstrated the effectiveness of HNN for a variety of downstream hypergraph representation learning tasks including hyperedge prediction and node classification where it achieved an overall mean gain in AUC of 7.72\% and 11.37\% across all models and hypergraphs, respectively. Finally, this work also demonstrated the utility of HNN for HTML style recommendation tasks and we make accessible our heterogeneous hypergraph benchmark for others to use.

\balance
\bibliographystyle{ACM-Reference-Format}
\bibliography{paper}

\clearpage
\pagebreak
\newpage

\appendix
\section*{Appendix} 

\section{Datasets}
A summary of the datasets used in Sec.~\ref{sec:exp} can be found in Table~\ref{table:dataset-stats-public}.

\begin{table}[h!]
\centering
\caption{Dataset statistics}
\label{table:dataset-stats-public}
\vspace{-2mm}
\begin{tabular}{l cccc}
\toprule
\textbf{Dataset} &  \textbf{\#Classes} &  \textbf{\#Nodes} &  \textbf{\#HyperEdges} & \textbf{\#Features} \\
\midrule
Cora-CA        & 7     & 2,708     & 1,072 &   1,433\\
DBLP & 6 & 43,413 & 22,535 &  1,425 \\
\midrule
Citeseer    & 6     & 3,327     & 4,732  &   3,703 \\
Cora-CC        & 7     & 2,708     & 1,579   &  1,433 \\
PubMed      & 3     & 19,717    & 7,963 & 500 \\
\bottomrule
\end{tabular}
\end{table}

\section{Hyperedge Prediction Difficulty}
\begin{figure}[h]
\centering
\includegraphics[width=0.95\linewidth]{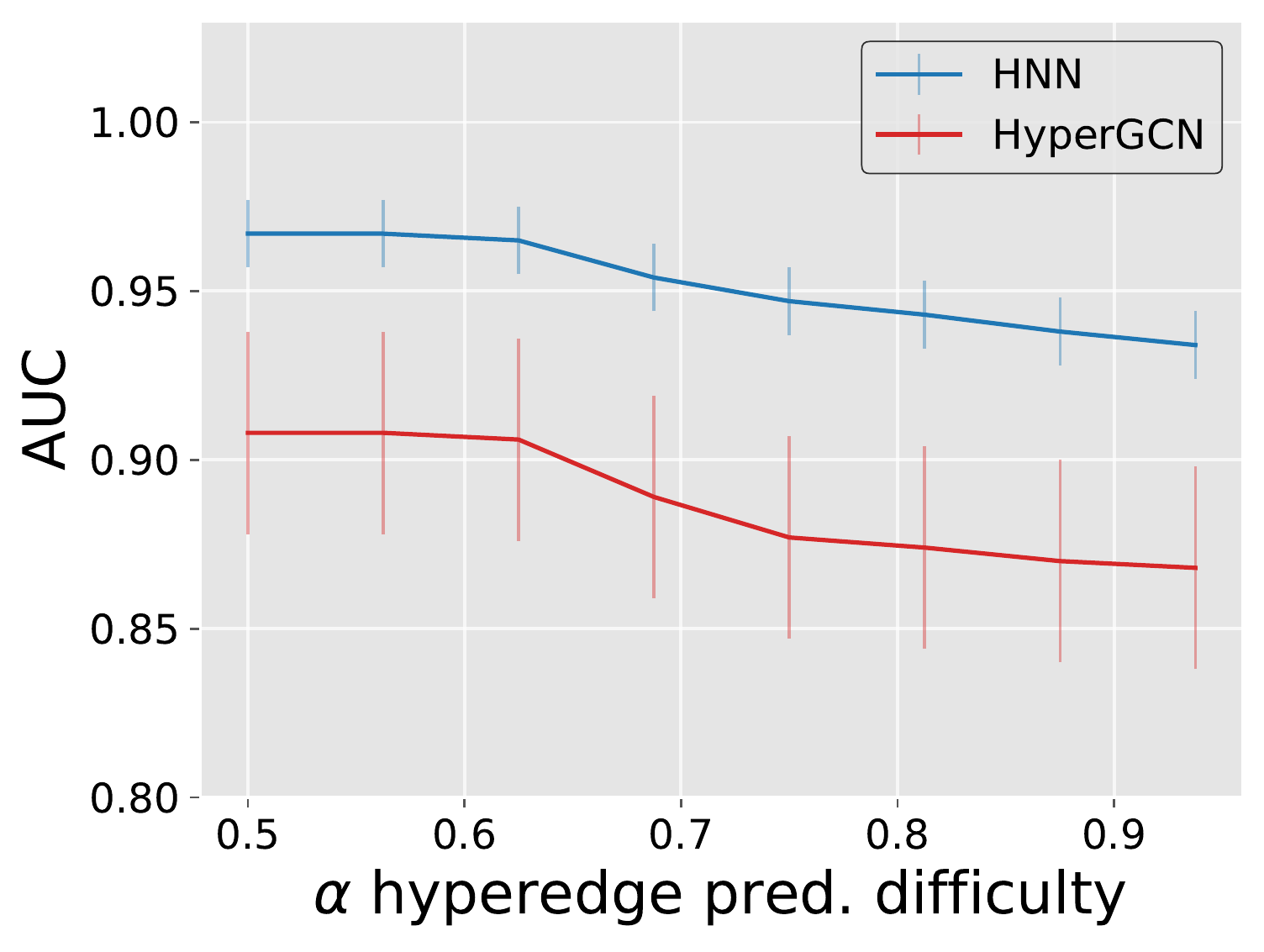}
\vspace{-2mm}
\caption{
Varying the difficulty of the hyperedge prediction task. 
For this experiment, as $\alpha$ increases, the negative hyperedges become increasingly similar to the observed hyperedges, therefore, increasing the difficulty of the 
task.
}
\label{fig:hyperedge-pred-diff}
\end{figure}

We also investigated increasing the difficulty of the hyperedge prediction task.
In Figure~\ref{fig:hyperedge-pred-diff}, we compare the best baseline model to HNN as the difficulty of the hyperedge prediction task increases.
Notably, as $\alpha$ increases, the negative hyperedges become increasingly similar to the observed hyperedges, therefore, increasing the difficulty of the hyperedge prediction task for Citeseer.
Most importantly, HNN achieves significantly better performance across all $\alpha$ while also having lower standard error indicating that HNN is more robust as the difficulty of the hyperedge prediction task increases.
As an aside, many additional results were removed due to space, however we observed similar findings.

\section{Hyperedge Prediction Loss}
In addition, we also investigated using a different scoring function for hyperedge prediction in Table~\ref{tab:hyperedge-prediction-results-vary-scoring-func}, that is, instead of Eq.~\ref{eq:hyperedge-score-func}, we used Eq.~\ref{eq:diff-max-min-eval-metric} to derive a score $f(e_t)$ of a hyperedge $e_t$ by taking the difference between its maximum and minimum value over all node embeddings in the hyperedge $e_t$.
Overall, we find that HNN also achieves the best predictive performance across all other models and across all the graphs as shown in Table~\ref{tab:hyperedge-prediction-results-vary-scoring-func}.
Notably, even when using a completely different loss function for hyperedge prediction, HNN still achieves the best performance across all models and graphs, indicating its overall utility is not tied to a specific loss function, graph, or even application task.
This result is also strong as it demonstrates the robustness of the HNN framework for leveraging a variety of different objective functions.

\section{Varying Learning Rate}
\begin{figure}[h]
\vspace{-2mm}
\centering
\subfigure{
\includegraphics[width=0.46\linewidth]{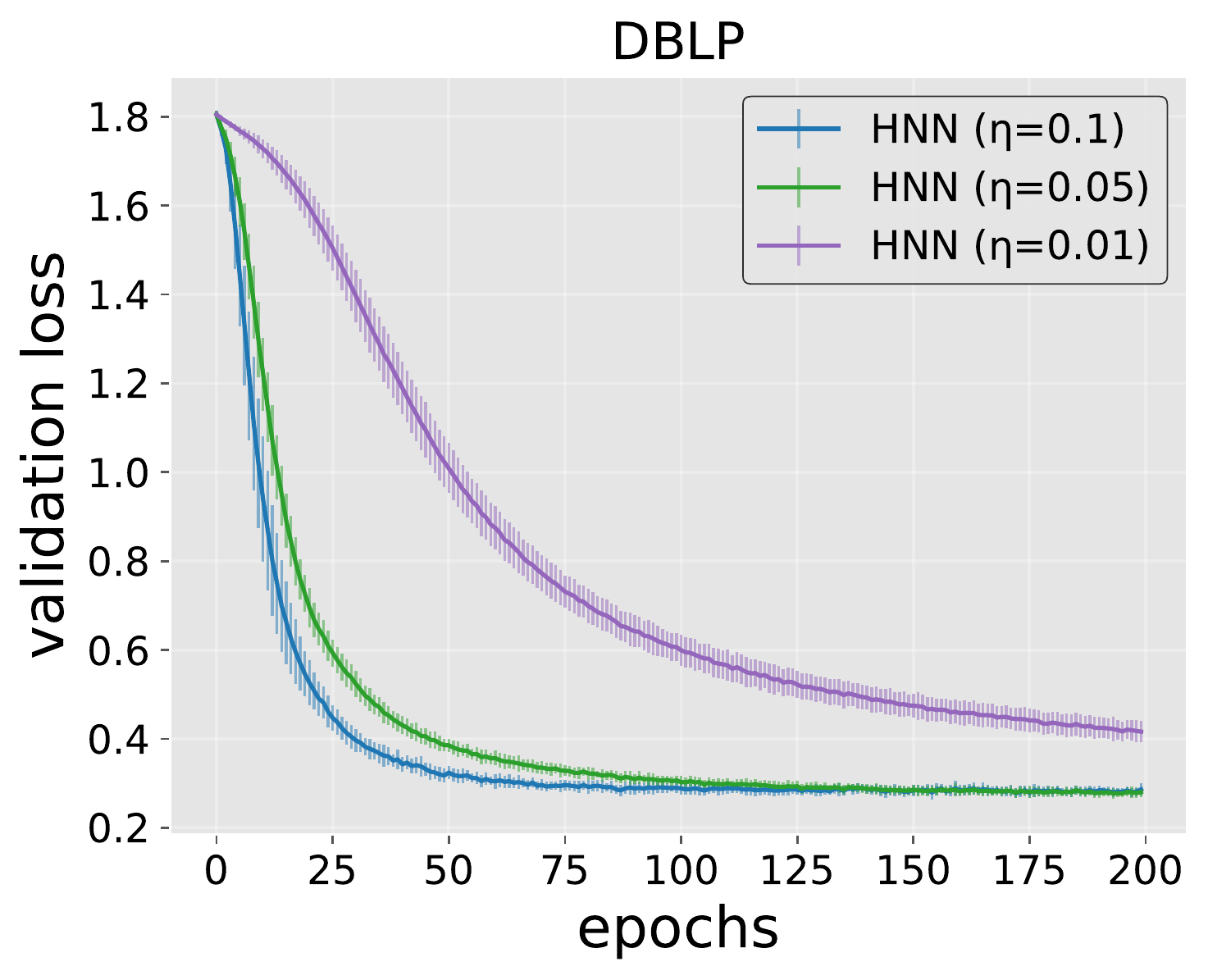}
\label{fig:loss-curves-vary-learning-rate-DBLP}
}
\subfigure{
\includegraphics[width=0.46\linewidth]{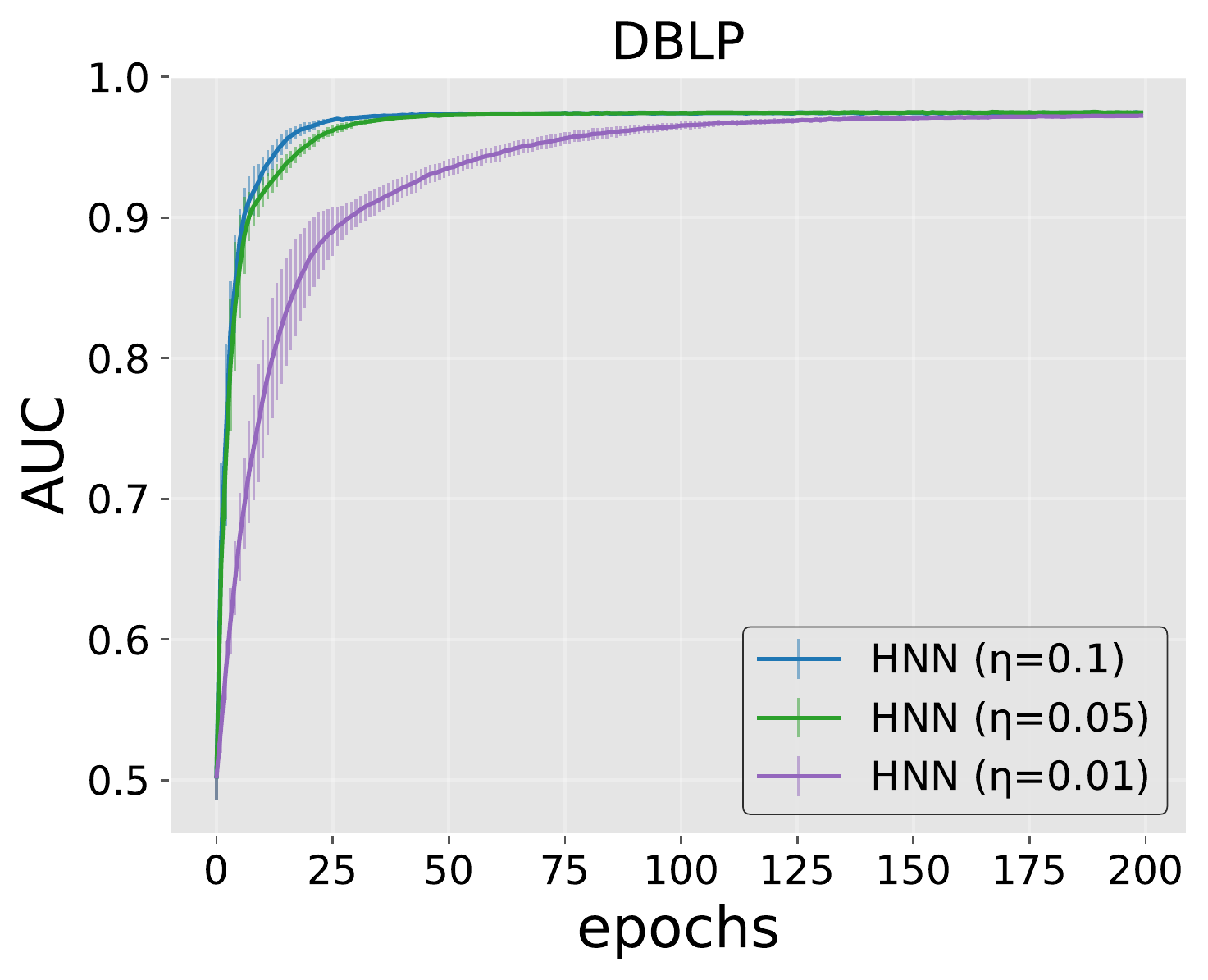}
\label{fig:AUC-curves-vary-learning-rate-DBLP}
}
\vspace{-4mm}
\caption{
Validation loss and AUC curves for HNN models trained by varying the learning rate $\eta \in \{0.1,0.05,0.01\}$.
}
\label{fig:loss-curves-vary-learning-rate}
\end{figure}

In Figure~\ref{fig:loss-curves-vary-learning-rate}, we vary the learning rate of HNN. 
Overall, we see that HNN with $\eta=0.1$ achieves the best AUC when using a small amount of epochs for training. 
However, as the number of epochs increases towards 200, all models converge to nearly the same AUC.

\section{Varying Layers and Epochs}
\begin{figure}[b]
\vspace{-2mm}
\centering
\includegraphics[width=0.80\linewidth]{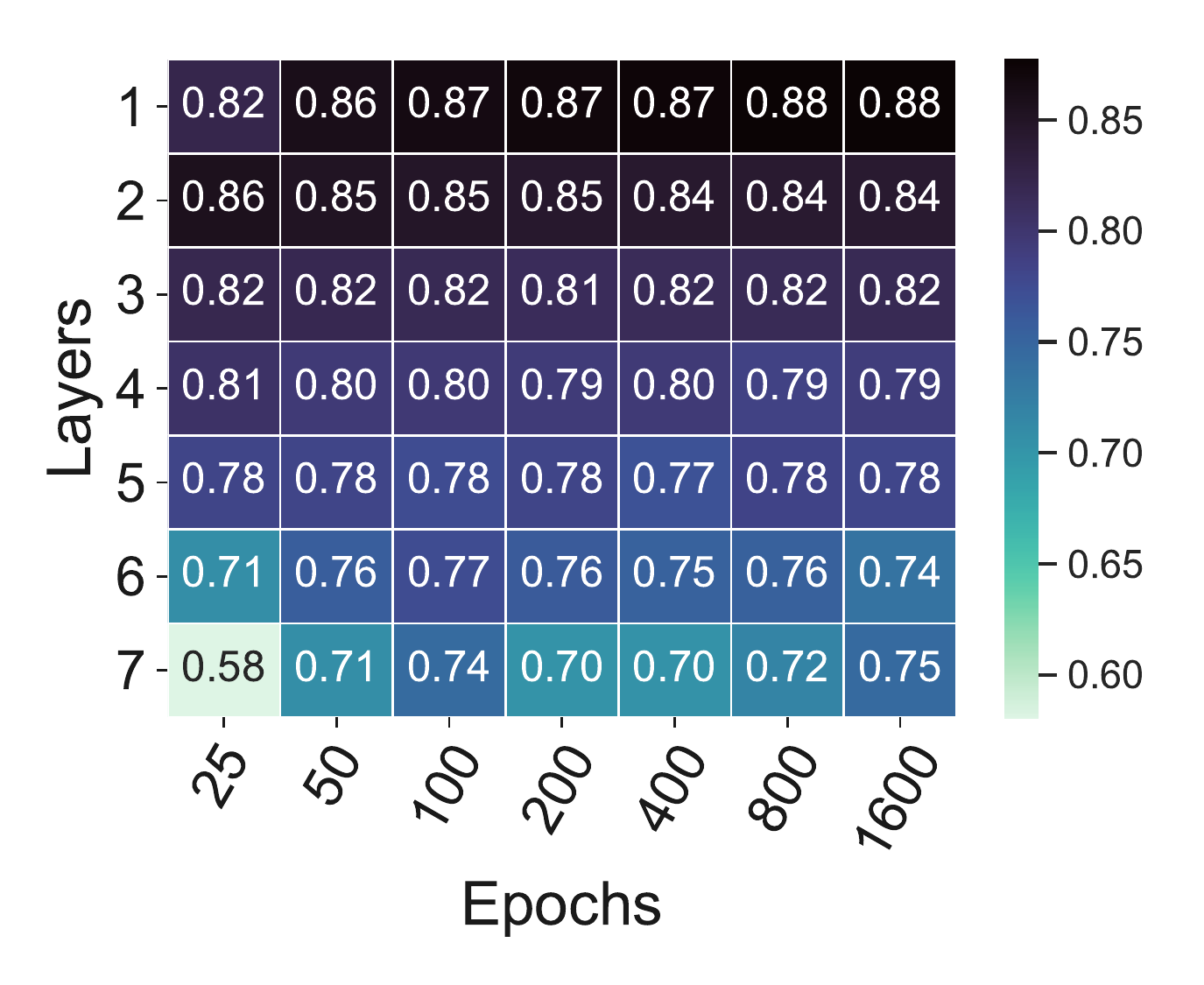}
\vspace{-4mm}
\caption{Varying number of epochs and layers.}
\label{fig:ablation-varying-epochs-vs-layers}
\end{figure}

In addition, Figure~\ref{fig:ablation-varying-epochs-vs-layers} shows the performance of HNN when the number of layers and epochs varies for citeseer.
Notably, AUC generally decreases as the number of layers become larger.
The best performance is achieved when HNN is configured to use a modest number of layers ($L=1$ or $2$) trained with a sufficient number of epochs.
For $L=1$, the best performance is achieved when trained with 800 or 1600 epochs.
We also observe that when HNN is trained with a few epochs such as 25, then HNN with $L=2$ achieves better predictive performance than the single layer model.
Hence, better performance can be achieved in this setting with more layers.

\begin{table*}[h]
\centering
\caption{Hyperedge prediction using a different hyperedge scoring function (Eq.~\ref{eq:diff-max-min-eval-metric}) for prediction of the hyperedges.}
\label{tab:hyperedge-prediction-results-vary-scoring-func}
\vspace{-2mm}
\begin{tabular}{l ccccc}
\toprule
& \multicolumn{1}{c}{\textbf{Citeseer}} & \multicolumn{1}{c}{\textbf{Cora-CC}} & \multicolumn{1}{c}{\textbf{PubMed}} & \multicolumn{1}{c}{\textbf{DBLP}} &
\multicolumn{1}{c}{\textbf{Cora-CA}} \\
\midrule

\textit{GCN} & 
0.881 $\pm$ 0.03 & 
0.869 $\pm$ 0.03 & 
0.761 $\pm$ 0.02 & 
0.916 $\pm$ 0.02 & 
0.882 $\pm$ 0.03 
\\

\textit{GraphSAGE} & 
0.903 $\pm$ 0.02 &
0.859 $\pm$ 0.02 & 
0.797 $\pm$ 0.02 & 
0.910 $\pm$ 0.02 & 
0.866 $\pm$ 0.05 
\\

\textit{HyperGCN-Fast} & 
0.839 $\pm$ 0.02 & 
0.831 $\pm$ 0.01 & 
0.793 $\pm$ 0.01 & 
0.844 $\pm$ 0.00 & 
0.783 $\pm$ 0.02 
\\

\textit{HyperGCN} & 
0.881 $\pm$ 0.01 &  
0.867 $\pm$ 0.02 & 
0.816 $\pm$ 0.02 & 
0.909 $\pm$ 0.01 & 
0.785 $\pm$ 0.05 
\\

\textit{HGNN} & 
0.811 $\pm$ 0.03 &
0.833 $\pm$ 0.01 & 
0.790 $\pm$ 0.01 & 
0.897 $\pm$ 0.00 & 
0.890 $\pm$ 0.03 
\\

\midrule

\textit{HNN} & 
\textbf{0.925 $\pm$ 0.02} &
\textbf{0.890 $\pm$ 0.01} & 
\textbf{0.864 $\pm$ 0.03} & 
\textbf{0.926 $\pm$ 0.03} & 
\textbf{0.894 $\pm$ 0.04} 
\\

\bottomrule
\end{tabular}
\end{table*}

\begin{table}[h]
\caption{Results for Button Style Recommendation.}
\label{table:results-style-button-nDCG}
\vspace{-2mm}
\small
\begin{tabular}{l cccc}
\toprule
&
\multicolumn{4}{c}{nDCG@K} 
\\ 
\cmidrule(lr){2-5} 
\textbf{Model} & 
@1 & @10 & @25 & @50 
\\ \midrule
Random &  
0.000 $\pm$ 0.00& 0.000 $\pm$ 0.00& 0.004 $\pm$ 0.00& 0.006 $\pm$ 0.00\\ 

Pop. & 
0.000 $\pm$ 0.00 & 0.002 $\pm$ 0.00& 0.003 $\pm$ 0.00& 0.005 $\pm$ 0.00\\ 

HyperGCN 
& 0.008 $\pm$ 0.01
& 0.009 $\pm$ 0.00
& 0.017 $\pm$ 0.01
& 0.021 $\pm$ 0.01
\\ 
\midrule

HNN 
& 0.243 $\pm$ 0.05
& 0.354 $\pm$ 0.04
& 0.368 $\pm$ 0.04
& 0.379 $\pm$ 0.04\\ 

\bottomrule
\end{tabular}
\end{table}

\begin{table}[h]
\caption{Results for Background Style Recommendation.}
\label{table:results-style-background-nDCG}
\vspace{-2mm}
\small
\begin{tabular}{l cccc}
\toprule
&
\multicolumn{4}{c}{nDCG@K} 
\\ 
\cmidrule(lr){2-5} 
\textbf{Model} & 
@1 & @10 & @25 & @50 
\\ \midrule
Random & 
0.000 $\pm$ 0.00 & 0.000 $\pm$ 0.00& 0.000 $\pm$ 0.00& 0.013 $\pm$ 0.00\\ 

Pop. & 
0.001 $\pm$ 0.00 & 0.006 $\pm$ 0.00& 0.010 $\pm$ 0.00& 0.016 $\pm$ 0.00\\ 

HyperGCN & 
0.000 $\pm$ 0.00&  0.018 $\pm$ 0.02& 0.025 $\pm$ 0.03& 0.041 $\pm$ 0.04\\ 

\midrule

HNN & 
0.181 $\pm$ 0.11 & 0.308 $\pm$ 0.11& 0.333 $\pm$ 0.11& 0.369 $\pm$ 0.10\\ 

\bottomrule
\end{tabular}
\end{table}

\section{Additional Style Rec. Results}
\begin{figure}[b]
\centering
\includegraphics[width=0.99\linewidth]{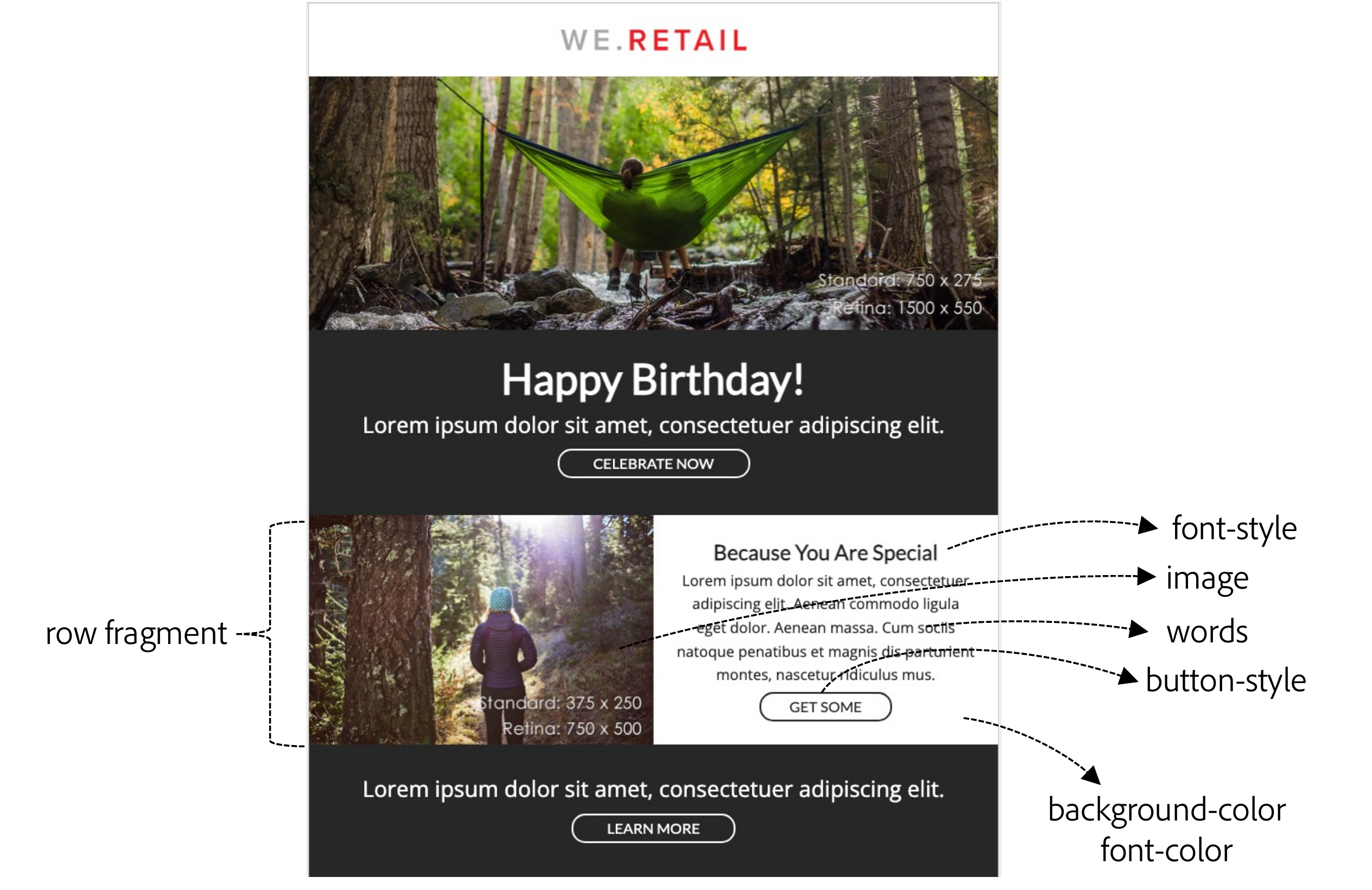}
\caption{Overview of a fragment and extracted entities.}
\label{fig:extraction-example}
\end{figure}

We also used the common evaluation metric of nDCG for evaluating HNN for style recommendation.
Results are provided in Table~\ref{table:results-style-button-nDCG}-\ref{table:results-style-background-nDCG} for button-style and background-style recommendation, respectively.
In all cases, HNN outperforms the other approaches across all K.
This result is consistent with the previous findings when using HR as discussed in Section~\ref{sec:email-style-recommendation}.
A summary of the entity types extracted and examples of each is provided in Table~\ref{table:entity-type-summary}. 
In Figure~\ref{fig:extraction-example}, we also provide an intuitive example of one of the fragments from a document in the corpus along with the different types of entities extracted from it.

\begin{table}[h]
\centering
\caption{Summary of entity (node) types extracted.}
\vspace{-2.5mm}
\label{table:entity-type-summary}
\renewcommand{\arraystretch}{1.10} 
\small
\begin{tabularx}{1.0\linewidth}{l l}
\toprule
\textbf{Node Type} &   \textbf{Example} \\
\midrule

\textsf{{button-style}} & 
\texttt{bg-color:\#3867FF; border-radius:50px;} \\
& \texttt{color:\#FFF; font-size: 16px;}\\

\textsf{text-style} & \texttt{font-size:12px; color:\#7C7C7C;} \\
\textsf{background \& font color} & \texttt{bg-color:\#DB2100; color:\#FFF;} \\
\textsf{background style} & \texttt{bg-color:\#fafafa; opacity:0.4;} \\
\textsf{words} & \texttt{connected}, \texttt{enjoy}, \texttt{favorite}, ... \\
\textsf{image} &  actual image \\
\textsf{entire fragment} & \texttt{$-$} \\
\bottomrule
\end{tabularx}
\end{table}

\end{document}